\documentclass{article}

\usepackage{arxiv}

\usepackage[utf8]{inputenc} 
\usepackage[T1]{fontenc}    
\usepackage{hyperref}       
\usepackage{url}            
\usepackage{booktabs}       
\usepackage{amsfonts}       
\usepackage{nicefrac}       
\usepackage{microtype}      
\usepackage{lipsum}		
\usepackage{graphicx}
\usepackage{natbib}
\usepackage{doi}

\usepackage{float}

\usepackage{multirow}
\usepackage{booktabs}
\usepackage{amsmath}
\usepackage{amssymb}

\usepackage{geometry}
\title{SylloBio-NLI: Evaluating Large Language Models on \\Biomedical Syllogistic Reasoning}

\author{
  \textbf{Magdalena Wysocka\textsuperscript{1,2}},
  \textbf{Danilo Carvalho\textsuperscript{1,2}},
  \textbf{Oskar Wysocki\textsuperscript{1,3}},
\\
  \textbf{Marco Valentino\textsuperscript{3}},
  \textbf{Andr\'e Freitas\textsuperscript{1,2,3}}
\\
\\
  \textsuperscript{1}Department of Computer Science, University of Manchester, United Kingdom\\
  \textsuperscript{2}National Biomarker Centre (NBC), CRUK Manchester Institute, United Kingdom\\
  \textsuperscript{3}Idiap Research Institute, Martigny, Switzerland
\\
  \small{
    \textbf{Correspondence:} {firstname.lastname}@manchester.ac.uk\textsuperscript{1}{firstname.lastname}@idiap.ch\textsuperscript{3}
  }
}

\renewcommand{\shorttitle}{SylloBio-NLI}

\hypersetup{
pdftitle={A template for the arxiv style},
pdfsubject={q-bio.NC, q-bio.QM},
pdfauthor={David S.~Hippocampus, Elias D.~Striatum},
pdfkeywords={First keyword, Second keyword, More},
}

\begin{document}
\maketitle

\begin{abstract}
Syllogistic reasoning is crucial for Natural Language Inference (NLI). This capability is particularly significant in specialized domains such as biomedicine, where it can support automatic evidence interpretation and scientific discovery.
This paper presents SylloBio-NLI\footnote{code and dataset available at: \url{https://github.com/neuro-symbolic-ai/SylloBio-NLI}\label{fn:repo}}, a novel framework that leverages external ontologies to systematically instantiate diverse syllogistic arguments for biomedical NLI. We employ SylloBio-NLI to evaluate Large Language Models (LLMs) on identifying valid conclusions and extracting supporting evidence across 28 syllogistic schemes instantiated with human genome pathways. Extensive experiments reveal that biomedical syllogistic reasoning is particularly challenging for zero-shot LLMs, which achieve an average accuracy between 70\% on generalized modus ponens and 23\% on disjunctive syllogism. At the same time, we found that few-shot prompting can boost the performance of different LLMs, including Gemma (+14\%) and LLama-3 (+43\%). However, a deeper analysis shows that both techniques exhibit high sensitivity to superficial lexical variations, highlighting a dependency between reliability, models' architecture, and pre-training regime. Overall, our results indicate that, while in-context examples have the potential to elicit syllogistic reasoning in LLMs, existing models are still far from achieving the robustness and consistency required for safe biomedical NLI applications.
\end{abstract}


\section{Introduction}
\label{sec:introduction}
Syllogistic reasoning  -- i.e., the process of deriving valid conclusions from premises through the systematic application of abstract reasoning schemes -- is a fundamental type of inference for developing and evaluating Natural Language Inference (NLI) models that can reason over textual evidence at scale (\cite{MacCartney2009, wu-etal-2023-hence}). within specialised domains such as biomedicine, the ability to reason over natural language can have significant practical implications for supporting complex discourse interpretation, scientific discovery, and the development of downstream biomedical and clinical applications (\cite{jullien-etal-2023-nli4ct,jullien-etal-2023-semeval}), allowing for a set of formally defined patterns for controlled inference.

Moreover, determining the reliability and robustness of NLI models with regard to syllogistic reasoning is a crucial type of assessment, particularly in critical domains requiring safety guarantees (\cite{eisape-etal-2024-systematic,jullien-etal-2024-semeval}). This type of evaluation is further motivated by two key factors. First, syllogistic reasoning is prevalent in discourse at large and, therefore, it is expected that approaches based on Large Language Models (LLMs) are exposed to a variety of reasoning schemes during pre-training (\cite{wu-etal-2023-hence}). Second, a model that learns to perform syllogistic reasoning should intrinsically possess the ability to generalize to different domains, regardless of the specific world knowledge acquired during pre-training (\cite{kim2024mechanistic}). This is because the ability to perform syllogistic reasoning should be content-independent, that is, the ability to derive logically valid conclusions is only a function of the formal logical schemes and should be independent of its concrete instantiation.
However, despite the importance of syllogistic reasoning for NLI and the abundance of benchmarks involving commonsense knowledge (\cite{yu2023naturallanguagereasoningsurvey}), resources for assessing how systematic reasoning capabilities transfer to specialised domains are still scarce and require substantial expert-level annotation effort to guarantee quality and correctness (\cite{ZHAO2024122807,eisape-etal-2024-systematic,porada-etal-2022-pre,wysocka2024large}). 

This paper focuses on advancing the availability of resources for biomedical syllogistic reasoning along with our understanding of the capabilities of state-of-the-art NLI models. In particular, we propose SylloBio-NLI, a novel framework for automatically generating NLI resources for evaluating syllogistic reasoning within biomedical domains. Specifically, we demonstrate how external domain-specific resources such as ontologies and thesauri can be leveraged to instantiate a wide range of syllogistic schemes for biomedical NLI tasks, including textual entailment and evidence extraction.

\begin{figure}[t]
\centering
  \includegraphics[width=.55\columnwidth]{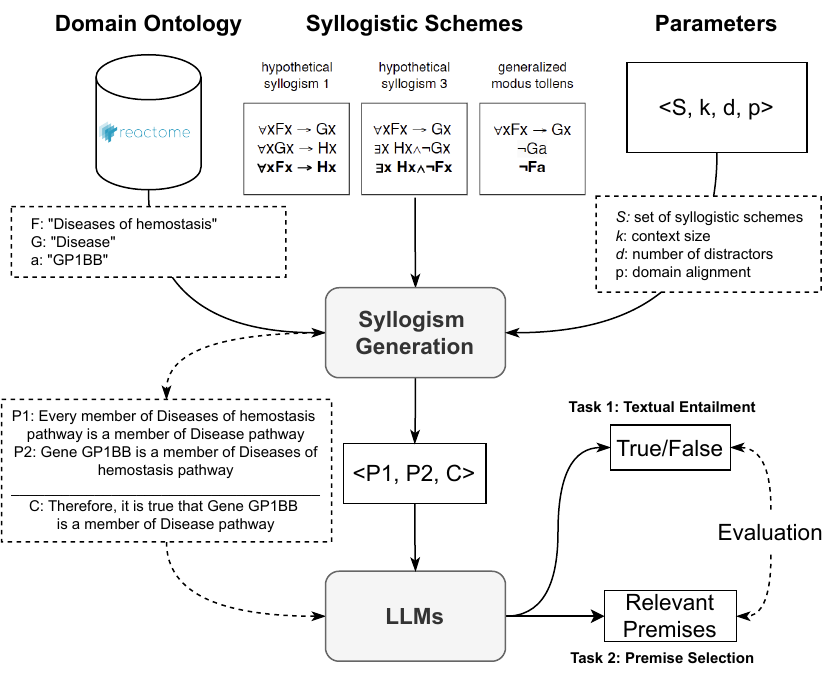}
  \caption{End-to-end diagram of the proposed methodological framework, illustrating the generation of syllogistic arguments from domain-specific ontologies, parameterized input to LLMs, and evaluation tasks including textual inference and premise selection.}
  \label{fig:framework}
\end{figure}

The methodological framework behind SylloBio-NLI is designed to address the annotation scarcity problem for the granular evaluation of complex reasoning in specialized domains. The framework minimises the human annotation effort while guaranteeing the correctness of the generated data by leveraging explicit domain knowledge in the ontologies and the systematicity of known syllogistic schemes (see Fig. \ref{fig:framework}).

By instantiating SylloBio-NLI on human biological pathways using Reactome (\cite{croft2014reactome}), we evaluate the domain-specific reasoning capabilities of 8 open-source LLMs on 28 syllogistic schemes, comparing the performance in a zero-shot (ZS) and few-shot (FS) settings. An extensive empirical evaluation led to the following findings and conclusions:

1. We determine that LLMs exhibit surprisingly low ZS performance on biomedical syllogistic arguments with an average accuracy between 70\% on \textit{generalised modus ponens} and 23\% on \textit{disjunctive syllogism} (where random performance is equal to 50\%). These low performances are generally shared across models' family (i.e., Llama-3, Mistral, Gemma, BioMistral) and pre-trained regimes (language modelling and instruction-tuning).

2. We found that the FS setting can improve the performance of different LLMs. In particular, we observe a significant boost for both Gemma and Llama-3, with an average increase in F1-score of 14\% and 43\% respectively. At the same time, the experiments reveal that such improvement is inconsistent across models' families and pre-training regimes.

3. We perform a robustness analysis adopting logically equivalent variations of the same syllogistic schemes by rephrasing the arguments via negations, complex predicates, and De Morgan's laws. Such analysis reveals that both ZS and FS techniques are highly sensitive to surface-form and lexical variations, demonstrating a shared inability to systematically abstract the underlying reasoning rules required to derive valid conclusions. These results indicate that, while FS has the potential to elicit syllogistic reasoning in LLMs, existing models are still far from achieving the robustness and consistency required for safe biomedical NLI. 

Overall, the above findings suggest that, upon granular inference scrutiny, the reasoning mechanisms induced in LLMs still confound formal and material inference patterns. Moreover, while there are FS intervention mechanisms which can improve models' performance, delivering controlled specialised syllogistic reasoning remains a challenge for LLMs at large.

To the best of our knowledge, this is the first work focusing on designing a methodology for evaluating syllogistic reasoning within specialised domains, thoroughly assessing the performance of LLMs, and releasing a domain-specific resource for supporting future work in the field. The code for the dataset generation and the evaluation pipeline is fully available online\footref{fn:repo}.


\section{SylloBio-NLI}
\label{sec:headings}

We introduce a general framework for developing resources to systematically evaluate the syllogistic reasoning capabilities of NLI models within biomedical domains.
In particular, SylloBio-NLI leverages the systematicity of known syllogistic schemes (\cite{Betz2021}) along with external thesauri/ontologies to scale up the generation of deductively valid arguments and to characterise inference performance across a granular set of syllogistic schemes. 

\begin{figure*}[t]
\centering
\includegraphics[width=.95\textwidth]{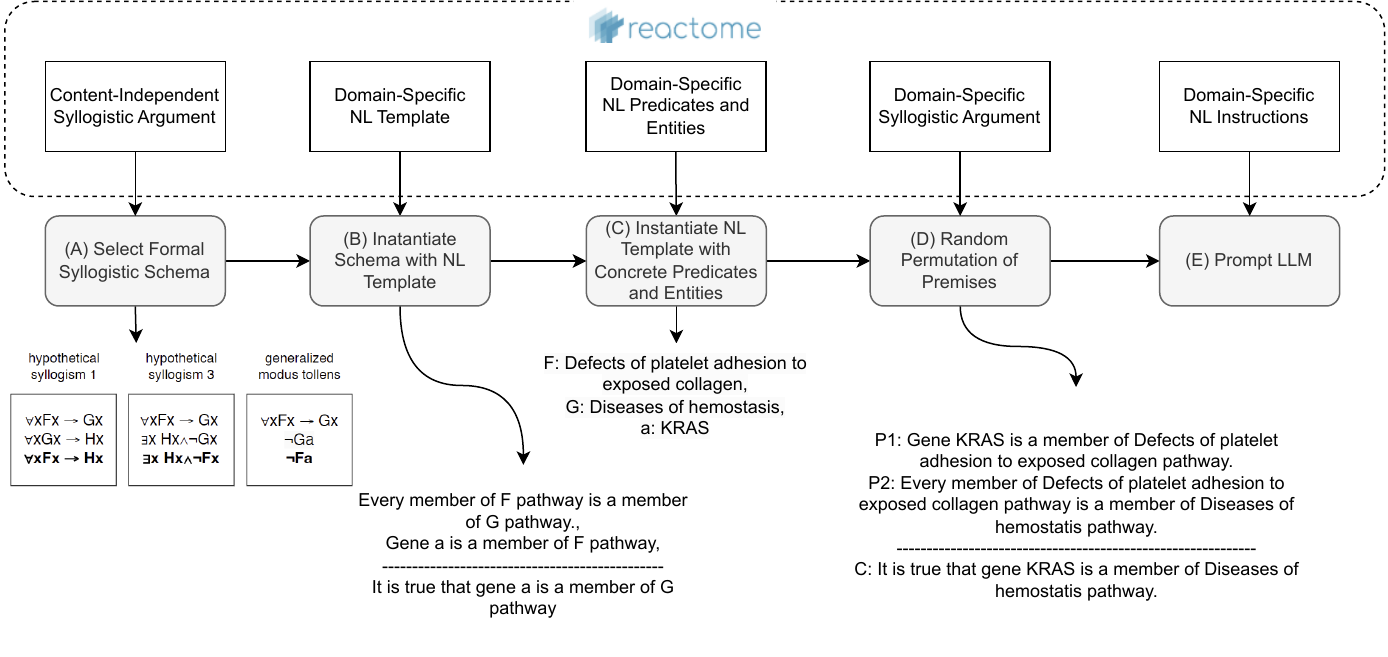}
\caption{Domain-specific pipeline for creating natural language instances of argument schemes with multiple templating. Steps include selecting a syllogistic schema (A), applying a domain-specific template (B), instantiating with predicates and entities (C), permuting premises (D), and prompting LLMs for evaluation (E).}
\label{fig:pipeline_domain-specific}
\end{figure*}

To this end, we focus on seven general syllogistic schemes, including \emph{generalized modus ponens}, \emph{generalized contraposition}, \emph{hypothetical syllogism 1}, \emph{hypothetical syllogism 3}, \emph{generalized modus tollens}, \emph{disjunctive syllogism} and \emph{generalized dilemma}. These base schemes are then adopted to generate 28 different variations of syllogistic argument templates applying negation, complex predicates, and De Morgan’s laws, which can then be instantiated with concrete domain knowledge (see Fig. \ref{fig:corpus_creation} in Appendix \ref{sec:appendix_pipeline}). The overall methodology is outline by the stages in Fig. \ref{fig:pipeline_domain-specific}.

First, for each syllogistic scheme, a corresponding formal argument scheme (consisting of abstract premises and conclusion expressed in first-order logic) is created (Fig. \ref{fig:pipeline_domain-specific}A). Next, each symbolic formula in the formal scheme is individually replaced by a natural language domain-specific sentence schema (Fig. \ref{fig:pipeline_domain-specific}B). For example, the formulae for the base schema of generalized modus ponens (i.e.,  Premise 1: $\forall xFx \Rightarrow Gx$, Premise 2: $Fa$, Conclusion: $Ga$) can be translated into the natural language template: \emph{``Premise 1: Every member of F is a member of G''}. Premise 2: \emph{``a is a member of F''}. Conclusion: \emph{``a is a member of G''}.

Subsequently, the entity and property placeholders in the natural language template are replaced argument-wise with domain-specific entities and predicates extracted from an external ontology (Fig. \ref{fig:pipeline_domain-specific}C).
In this process, the syllogistic schemes provide the logical validity component of the formal inference while the ontology subdomain and its associated mapping to the schemes provide the soundness (content-based) for the premises and conclusions. Hence, we obtain instances of syllogistic arguments in natural language that are stored in a knowledge base of premises and conclusions.

Finally, the natural language syllogistic arguments can be structured to create prompts for evaluating LLMs. This involves introducing the syllogistic argument, clearly framing the premises, and specifying the instructions for the inference, as illustrated in Appendix  \ref{sec:appendix_prompting_zs}.

\subsection{Biomedical NLI Tasks}

SylloBio-NLI takes advantage of the corpus's design, where every premise and conclusion in each argument is explicitly stated and fully transparent (Fig. \ref{fig:pipeline_domain-specific}). Once the full set of natural language syllogistic arguments is generated, the premises and conclusions in each instance can be adopted to define two different NLI tasks:

\paragraph{Task 1.}  A \emph{textual entailment} task where the model determines whether the conclusion logically follows from the premises, with the output being 'True' or 'False'. 
The task is designed using natural language syllogisms generated from formal schemes, with valid and invalid argument instances. The model evaluates each argument's logical validity, requiring it to reason deductively based on the given premises. The focus is on testing the ability to discern valid reasoning structures in a biomedical context, independent of factual correctness.

Formally, given a set of premises \( \mathcal{P} = \{ P_1, P_2, \dots, P_n \} \) and a conclusion \( C \), the task is to determine whether the logical entailment holds: 

$$\mathcal{P} \models C$$
   
such that (i) The premises and conclusion are instantiated from formal syllogistic schemes and (ii) arguments may be \emph{valid} or \emph{invalid}.  The assessment is purely on logical validity, independent of the factual correctness of premises and conclusion.

\paragraph{Premises:} A set \( \mathcal{P} = \{ P_1, P_2, \dots, P_n \} \) of premises, where each \( P_i \) is a natural language sentence derived from syllogistic schemes.

\paragraph{Conclusion:} A natural language sentence \( C \), representing the conclusion.

\paragraph{Logical Validity Indicator:} A binary output \( \text{Output} \in \{ \text{True}, \text{False} \} \) such that:
    \[
    \text{Output} =
    \begin{cases}
        \text{True} & \text{if } \mathcal{P} \models C, \\
        \text{False} & \text{if } \mathcal{P} \not\models C.
    \end{cases}
    \]
    Here, \( \mathcal{P} \models C \) denotes that \( C \) is a logical consequence of \( \mathcal{P} \).

\paragraph{Task 2.} A \emph{premise selection} task where the model has to identify which premises are necessary and sufficient to justify the conclusion from Task 1, noting that some premises may be irrelevant. 
The task involves presenting the model with a set of premises, including both relevant and distractor premises. The model must correctly select the subset of premises that logically support the conclusion, excluding those that do not contribute to the entailment. This tests the ability to filter essential information and assess the understanding of the logical relationships between premises and conclusions.
Task 2 can be formalized as follows: 

\paragraph{Objective:} Identify the minimal subset of premises \( \mathcal{P}' \subseteq \mathcal{P} \) that are necessary and sufficient to logically derive the conclusion \( C \).

\begin{itemize}
    \item \textbf{Given:}
    \begin{itemize}
        \item A set of premises \( \mathcal{P} = \{ P_1, P_2, \dots, P_n \} \).
        \item A conclusion \( C \).
    \end{itemize}
    \item \textbf{Task Definition:} Find the minimal subset \( \mathcal{P}' \subseteq \mathcal{P} \) satisfying:
    \begin{enumerate}
        \item \textbf{Logical Sufficiency:}
        \[
        \mathcal{P}' \models C.
        \]
        \item \textbf{Minimality:}
        \[
        \nexists \ \mathcal{P}'' \subsetneq \mathcal{P}', \ \text{such that } \mathcal{P}'' \models C.
        \]
        \item \textbf{Necessity:} Every premise in \( \mathcal{P}' \) is necessary for the entailment.
    \end{enumerate}
\end{itemize}

\paragraph{Premises:} A set \( \mathcal{P} = \{ P_1, P_2, \dots, P_n \} \) containing both relevant and irrelevant (distractor) premises.

\paragraph{Conclusion:} A natural language sentence \( C \).

\begin{itemize}
    \item \textbf{Selected Premises:} A subset \( \mathcal{P}' \subseteq \mathcal{P} \) such that:
    \[
    \mathcal{P}' \models C,
    \]
    and \( \mathcal{P}' \) is minimal and necessary.
\end{itemize}.

\subsection{Human Genome Pathways for Syllogistic Reasoning Evaluation}

To instantiate the SylloBio-NLI methodology, we developed a specialized dataset using Reactome\footnote{\url{https://reactome.org}\label{fn:reactome}}, a comprehensive knowledge base containing detailed information on human biological pathways, including gene functions and interactions.  

Reactome's hierarchical structure, with its well-defined hierarchical gene-pathway membership relations, enables a systematic instantiation of the syllogistic arguments. This allows for the efficient generation of domain-specific NLI tasks that assess LLMs' ability to reason about biological pathway membership, controlling for different levels within the hierarchy. This use case is widely relevant in the context of interpreting pathway-level interactions, disease and treatment response mechanisms at a genomics level (\cite{fang2019discovering}).

By focusing on the Disease super-pathway, the largest and most intricate group within Reactome, we generated a diverse set of syllogistic schemes (see Fig. \ref{fig:pipeline_domain-specific}) that mirror the complex molecular interactions and regulatory mechanisms underlying disease processes. This approach ensures our evaluation reflects the depth and specificity needed to rigorously test the syllogistic reasoning properties of NLI models in this highly specialised and clinically significant domain. A comprehensive and systematic assessment of these properties is essential if LLMs are to be applied to this domain.

The resulting dataset includes 12,098 entity-gene names, which are used as substitutes for \textit{entity} placeholders. In contrast, pathway names, derived from different levels of biological hierarchies, are treated as \textit{predicate} placeholders because they describe relationships or actions involving the entities, rather than being entities themselves. The corpus includes 3767 complex predicates in total. The premises of the natural language argument are randomly re-ordered to mitigate potential biases during the evaluation (Fig. \ref{fig:pipeline_domain-specific}D).

\subsection{Dictionary of Gene and Pathway Membership}
\label{sec:appendix_dictionary}

Reactome\footref{fn:reactome} (version 88—March 2024) has entries for 11 226 protein-coding genes involved in 15 212 human reactions annotated from 38 549 literature references. These reactions are grouped into 2 698 pathways collected under 29 superpathways (e.g. Immune System) that describe normal cellular functions. Each superpathway is represented as a roughly circular ‘burst,’ with the central node corresponding to the top-level of the Reactome event hierarchy and concentric rings representing increasingly specific levels of the event hierarchy (sub-pathways) (e.g. Disease → Diseases of signal transduction by growth factor receptors and second messengers → Signalling by EGFR in Cancer → Signalling by Ligand-Responsive EGFR Variants in Cancer → Constitutive Signalling by Ligand-Responsive EGFR Cancer Variants). The relationships between these pathways are captured through parent-child arcs, reflecting the ontological "is-a" relationships.

The 29 Reactome superpathways group are each organized as a roughly circular ‘burst’. However, we built the corpus based on one, largest group of pathways called Disease. The central node of the Disease burst corresponds to the uppermost level of the Reactome event hierarchy (Table \ref{tab:unique_entities}). Concentric rings of nodes around the central node represent successive more specific levels of the event hierarchy (e.g. Disease → Diseases of signal transduction by growth factor receptors and second messengers → Signalling by EGFR in Cancer → Signalling by Ligand-Responsive EGFR Variants in Cancer → Constitutive Signalling by Ligand-Responsive EGFR Cancer Variants). The arcs connecting nodes between successive rings within a burst represent parent–child (is-a) relationships in the event hierarchy. When a specific pathway is shared by more than one burst, arcs connect its nodes between bursts. A node's size is proportional to the number of physical entities (proteins, complexes, chemicals) it contains.

\begin{table}[H]
  \centering
\resizebox{.48\textwidth}{!}{%
  \begin{tabular}{lcc}
    \hline
    \textbf{Level of pathway} & \textbf{\begin{tabular}[c]{@{}l@{}}Nr of unique \\ pathway names\\ \end{tabular}} & \textbf{\begin{tabular}[c]{@{}l@{}}Nr of unique \\ genes names\\ \end{tabular}}  \\
    \hline
    top-level      &          1 &                           \\
    sub-pathway     &        13 &           2131                \\
    sub-sub-pathway       &      59     &    2131                       \\
    sub-sub-sub-pathway &      345 &             2114              \\
    sub-sub-sub-sub-pathway    &  1110                       &     1775     \\
    
    sub-sub-sub-sub-sub-pathway    &   994                      &       1002    \\\hline
  \end{tabular}
  }
  \caption{\label{tab:unique_entities}
    Summary of a dictionary with the true taxonomic relationships between pathways and genes.
  }
\end{table}

\subsection{Reasoning Challenges}

\paragraph{Domain-specific Reasoning. }

Biomedical datasets differ from general datasets in that they involve complex semantic structures, such as gene-pathway relationships and hierarchies, where sentences encode intricate biological interactions and dependencies. Reasoning about statements such as "\textit{COL1A1 is involved in the TGF-beta signalling pathway}" requires not only an understanding of specific gene functions but also the ability to navigate over hierarchical biological pathways and their associated relations. This demands a higher level of domain-specific reasoning and the capability to controllably interpret relationships within a highly specialised context.

\paragraph{Material Inference Component.} Each sentence in the formal scheme (premises and conclusion) is linked to the domain knowledge, i.e. the membership of genes to a given level of pathway, which refers to a multi-level, hierarchy. For example, if gene X is a member of pathway Y, and that pathway is a child of a top-level pathway Z, the model must be able to infer that the gene also belongs to a top-level pathway.

\paragraph{Formal Inference Component.} Syllogisms are examples of content-independent formal inference in which, provided the truth value of the premises, the conclusion can be derived through the application of specific logical inference rules. This form of reasoning can pose challenges to models that are incapable of abstracting inference patterns from text and that are affected by content biases in their representation (\cite{prange2023challenges,kim2024mechanistic}). Moreover, the syllogistic schemes in this paper are designed to assess the interpretation of fine-grained logical operators including quantifiers, implications and negation, which have been proved to be particularly challenging for NLI models (\cite{pitler2023logic}).

\section{Empirical Evaluation}

\subsection{Model Architectures}

We used the proposed methodology and resources to assess the syllogistic NLI inference properties of eight open-source LLMs. 

Specifically, We test a range of architectures, including mistralai/Mistral-7B-v0.1, mistralai/Mistral-7B-Instruct-v0.2 (\cite{jiang2023mistral}), mistralai/Mixtral-8x7B-Instruct-v0.1 (\cite{mistral2023mixtral}), google/gemma-7b, google/gemma-7b-it (\cite{google2024gemma}), meta-llama/Meta-Llama-3-8B, meta-llama/Meta-Llama-3-8B-Instruct (\cite{llama3modelcard}), BioMistral/BioMistral-7B (\cite{labrak2024biomistral}).
Details of the models, access, parameters, and the prompts used are available in the Appendix \ref{sec:appendix_llms}, \ref{sec:appendix_experimental}, \ref{sec:appendix_prompting_zs}, \ref{sec:appendix_prompting_fs}.

\subsection{Evaluation Metrics}

Accuracy, F1-score, Recall, and Precision are used to evaluate performance in Task 1, a binary classification task where the conclusion \( C \) is labelled as either True or False. These metrics compare the predicted labels \( \hat{y}_i \) against the annotated gold labels \( y_i \) (True/False). 

We used Reasoning Accuracy $RA$ to evaluate the models' performance in Task 2 to correctly predict the entailment label \( \hat{y}_i \) and select the appropriate subset of premises \( \hat{P}_i \) that justify this prediction. The metric is calculated as the percentage of responses where both the predicted label \( \hat{y}_i \) matches the ground truth \( y_i \) and the selected premises \( \hat{P}_i \) match the ground truth premises \( P_i \). In addition, we use a measure of faithfulness introduced by \citet{jullien-etal-2024-semeval}, which assesses the ability of a model to correctly change its prediction when altering the truth value of the conclusion (see Appendix \ref{sec:appendix_evaluation} for additional details).

\begin{table*}[t]
  \centering
  \resizebox{1.0\textwidth}{!}{%
  \begin{tabular}{lllllllll|llllllll}
    \toprule
    \textbf{Model} & \multicolumn{8}{c}{\textbf{ZS}} & \multicolumn{8}{c}{\textbf{FS}} \\
     & \begin{tabular}[c]{@{}l@{}}Non-empty \\ output\end{tabular} & \begin{tabular}[c]{@{}l@{}}Irrelevant \\ text\end{tabular} & \begin{tabular}[c]{@{}l@{}}Following \\ instruction\end{tabular} & \textbf{Acc.} & \textbf{Recall} & \textbf{Precision} & \textbf{F1} & \textbf{Faith.} & \begin{tabular}[c]{@{}l@{}}Non-empty \\ output\end{tabular} & \begin{tabular}[c]{@{}l@{}}Irrelevant \\ text\end{tabular} & \begin{tabular}[c]{@{}l@{}}Following \\ instruction\end{tabular} & \textbf{Acc.} & \textbf{Recall} & \textbf{Precision} & \textbf{F1} & \textbf{Faith.} \\
    \toprule
    BioMistral-7B & 0.00 & - & - & - & - & - & - & - & 0.00 & - & - & - & - & - & - & - \\ \midrule
    Meta-Llama-3-8B & 0.03 & 0.03 & 0.01 & 0.00 & 0.01 & 0.01 & 0.01 & 0.00 & 0.69 & 0.01 & 0.68 & 0.35 & 0.51 & 0.39 & 0.44 & 0.19 \\
    Meta-Llama-3-8B-Instruct & \textbf{1.00} & 0.86 & 0.14 & 0.11 & 0.18 & 0.16 & 0.17 & 0.04 & \textbf{1.00} & 0.99 & 0.01 & 0.00 & 0.00 & 0.00 & 0.00 & 0.00 \\ \midrule
    Mistral-7B-Instruct-v0.2 & \textbf{1.00} & 0.1 & 0.9 & 0.47 & 0.63 & 0.48 & 0.54 & 0.48 & \textbf{1.00} & 0.24 & 0.76 & 0.45 & 0.44 & 0.45 & 0.44 & \textbf{0.46} \\
    Mistral-7B-v0.1 & 0.88 & 0.88 & 0.00 & 0.00 & 0.00 & 0.00 & 0.00 & 0.00 & 0.69 & 0.69 & 0.00 & 0.00 & 0.00 & 0.00 & 0.00 & 0.00 \\
    Mixtral-8x7B-Instruct-v0.1 & \textbf{1.00} & 0.37 & 0.63 & 0.52 & 0.45 & 0.53 & 0.49 & 0.35 & \textbf{1.00} & 0.09 & 0.91 & \textbf{0.61} & 0.44 & \textbf{0.67} & 0.53 & \textbf{0.46} \\ \midrule
    Gemma-7b & 0.99 & 0.2 & 0.79 & 0.42 & \textbf{0.76} & 0.45 & 0.57 & 0.09 & \textbf{1.00} & 0.00 & \textbf{1.00} & 0.60 & \textbf{1.00} & 0.55 & \textbf{0.71} & 0.19 \\
    Gemma-7b-it & \textbf{1.00} & 0.00 & \textbf{1.00} & \textbf{0.64} & 0.71 & \textbf{0.62} & \textbf{0.66} & \textbf{0.66} & \textbf{1.00} & 0.03 & 0.97 & 0.63 & 0.46 & 0.69 & 0.55 & 0.42 \\
    \bottomrule\\
  \end{tabular}
  }
  \resizebox{0.75\columnwidth}{!}{%
    \begin{tabular}{@{}llllll|llllllll@{}}
    \toprule
    \textbf{Model} & \multicolumn{5}{c}{\textbf{ZS}} & \multicolumn{5}{c}{\textbf{FS}} \\
     & \begin{tabular}[c]{@{}l@{}}Non-empty \\ output\end{tabular} & \begin{tabular}[c]{@{}l@{}}Irrelevant \\ text\end{tabular} & \begin{tabular}[c]{@{}l@{}}Following \\ instruction\end{tabular} & \textbf{RA} & \textbf{Faith.} & \begin{tabular}[c]{@{}l@{}}Non-empty \\ output\end{tabular} & \begin{tabular}[c]{@{}l@{}}Irrelevant \\ text\end{tabular} & \begin{tabular}[c]{@{}l@{}}Following \\ instruction\end{tabular} & \textbf{RA} & \textbf{Faith.} \\
    \toprule
    BioMistral-7B &  0.00 & 0.00 & - & - &  -  &  0.00  &  -  &  -  &  -  &  -  \\ \midrule
    Meta-Llama-3-8B & 0.02 & 0.018 & 0.002 & 0.00 & 0.00 & 0.87 & 0.01 & 0.87 & 0.35 & 0.31 \\
    Meta-Llama-3-8B-Instruct & \textbf{1.00} & 0.36 & 0.64 & 0.30 & 0.33 & \textbf{1.00} & 0.99 & 0.01 & 0.01 & 0.00 \\ \midrule
    Mistral-7B-Instruct-v0.2 & \textbf{1.00} & 0.02 & \textbf{0.98} & 0.26 & 0.64 & \textbf{1.00} & 0.05 & 0.95 & 0.34 & 0.65 \\
    Mistral-7B-v0.1 & \textbf{1.00} & 1.00 & 0.00 & 0.00 & 0.00 & 0.84 & 0.84 & 0.00 & 0.00 & 0.00 \\
    Mixtral-8x7B-Instruct-v0.1 & \textbf{1.00} & 0.05 & 0.95 & 0.34 & 0.63 & \textbf{1.00} & 0.05 & 0.95 & 0.24 & 0.46 \\ \midrule
    Gemma-7b & 0.69 & 0.31 & 0.38 & 0.23 & 0.14 & \textbf{1.00} & 0.00 & \textbf{1.00} & \textbf{0.58} & \textbf{0.69} \\
    Gemma-7b-it & \textbf{1.00} & 0.03 & 0.97 & \textbf{0.55} & \textbf{0.65} & \textbf{1.00} & 0.00 & \textbf{1.00} & 0.55 & 0.31 \\
    \bottomrule
  \end{tabular}
  }
  \caption{Main results for Task 1 (i.e., textual entailment, top) and Task 2 (i.e., premise selection, bottom).
  }
  \label{tab:response_type_ndistr_0}
\end{table*}

\subsection{Main Results and Discussion}
\begin{figure*}[t]
  \centering
  \begin{tabular}{cc}
    \includegraphics[width=.45\columnwidth]{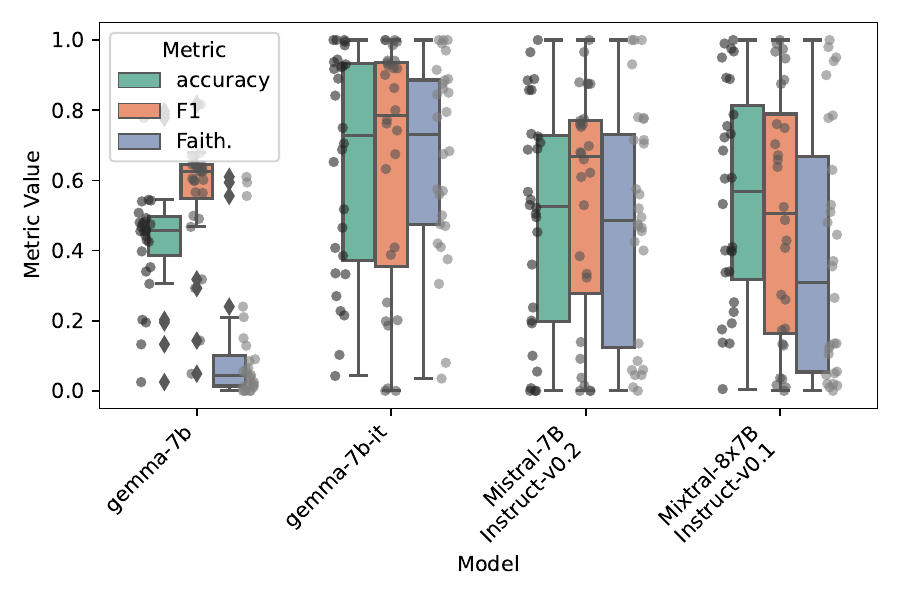} & 
    \includegraphics[width=.45\columnwidth]{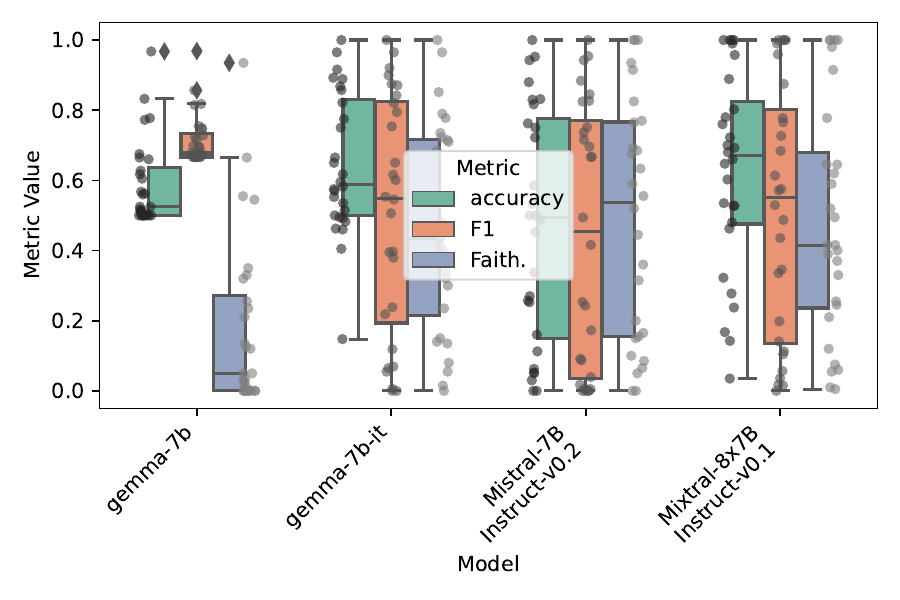} \\
  \end{tabular}
  \caption {Comparative Analysis of accuracy, F1, and Faithfulness across two prompt types: ZS (left) and FS (right) for Task 1 for the four best models.}
  \label{fig:faithfulness_task1}
\end{figure*}

\begin{figure*}[t]
\centering
\begin{tabular}{c}
    \includegraphics[width=\textwidth]{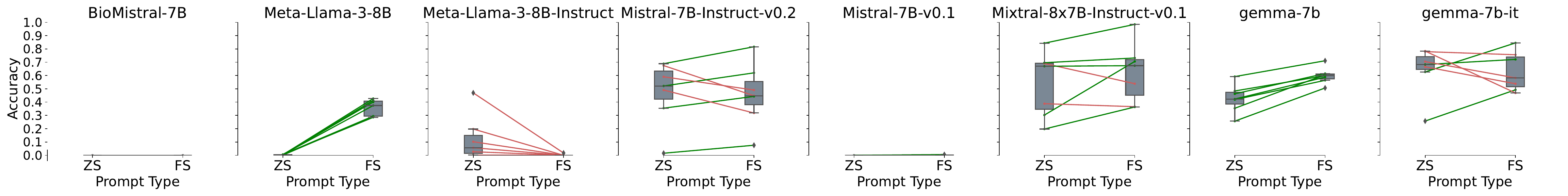} \\
    \includegraphics[width=\textwidth]{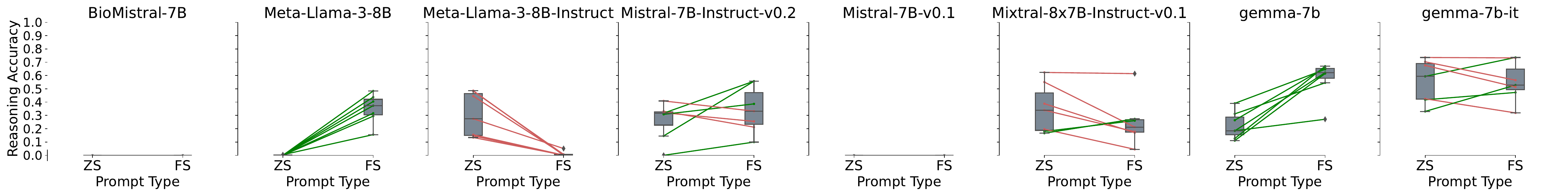} \\
\end{tabular}
\caption{Accuracy across two prompt types: ZS and FS, for Task 1 (top) and Task 2 (bottom). The lines connect the average accuracy for each of the seven syllogistic argument schemes, with green lines indicating an increase and red lines indicating a decrease. Gray boxplots display the median, Q1, Q3, and minimum and maximum values.}
\label{fig:combined_accuracy_by_promptstrategy_task12}
\end{figure*}

\begin{figure*}[t]
\centering
\begin{tabular}{c}
    \includegraphics[width=.95\textwidth]{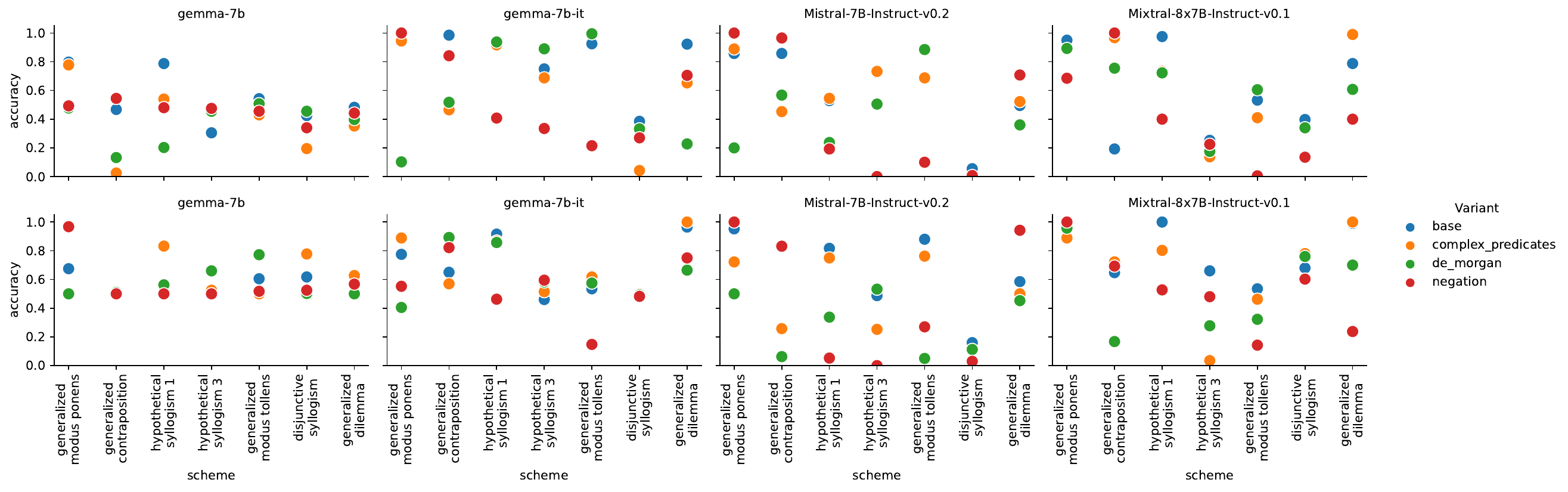} \\ \midrule
     \includegraphics[width=.95\textwidth]{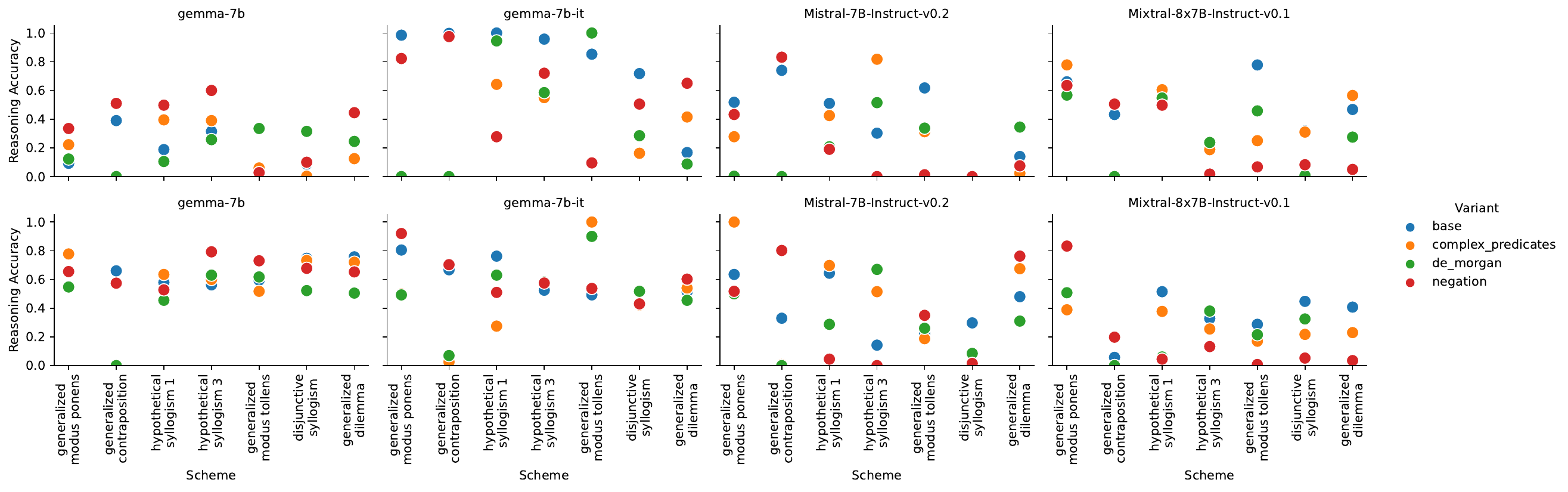} \\
\end{tabular}
\caption{Robustness to lexical variations. The graphs show the accuracy by model, syllogistic schema and lexical variations for ZS and FS respectively for Task 1 (top) and Task 2 (bottom).
}
\label{fig:task1_byvariant}
\end{figure*}

The main results on Task 1 and 2 are reported in Table \ref{tab:response_type_ndistr_0}. In particular, it is possible to derive the following main conclusions:

\paragraph{ZS LLMs struggle with controlled, domain-specific syllogistic inference.}

Overall, the results demonstrate that LLMs exhibit significant challenges when performing biomedical syllogistic reasoning in a ZS setting (Table \ref{tab:response_type_ndistr_0}).
We found, in fact, that the majority of the models struggle to achieve performances that are above the 50\% random accuracy on Task 1, with Gemma-7B-it being the only exception, achieving an accuracy of 64\%.
Similarly, in Task 2, Gemma-7B-it was again the best-performing model, achieving \textit{reasoning accuracy} of 55\%. In general, when comparing the performance on the two tasks, we observed that the models achieved lower performance in the premise selection task compared to textual entailment, indicating a shared inability of ZS models to identify the sufficient and necessary set of premises required for the entailment to hold.
In addition to low performance, we found a discrepancy between faithfulness and accuracy on both tasks, which highlights the inability to capture the underlying reasoning phenomena along with the presence of biases in the inference process. For example, we observed that for Gemma-7B, the ZS accuracy and F1-score are $\approx$ 0.5-0.6, while faithfulness plummets below 0.1 (Fig. \ref{fig:faithfulness_task1}). This is due to the tendency of the model to frequently output "True" (see Figs. \ref{fig:eight_plots_zeroshot}-\ref{fig:eight_plots_all_ICL} in Appendix \ref{sec:appendix_figures} and Appendix \ref{sec:appendix_results_responses}) regardless of the truth value of the conclusion. In contrast, we observed that such a phenomenon is less prominent in Gemma-7B-it where faithfulness is comparable to accuracy on both Task 1 (i.e., 0.66) and Task 2 (0.65). 
Overall, the results reveal that instruction-tuned models can obtain higher performances in the ZS setting, while models that are pre-trained on language modelling struggle to follow the provided natural language instructions. Moreover, we observed that this trend is independent of the specific knowledge the LLM is exposed to during pre-training. For instance, BioMistral-7B, the only biomedical domain-specific model, entirely failed to follow the instructions and generate outputs that are relevant to the target tasks.

\paragraph{FS can improve performance, but its impact is inconsistent across models' families and pre-training regimes.}

We found that the FS setting can improve the performance of different LLMs. In particular, we observe a significant boost for both Gemma-7b and Meta-Llama-3-8b, with an average increase in F1-score of 14\% and 43\% respectively. However, we observe such improvement to be inconsistent across models' family and pre-training regime, indicating variability in how different models handle prompt types in syllogistic reasoning tasks within the biomedical domain (Fig. \ref{fig:combined_accuracy_by_promptstrategy_task12}).
In particular, the results show that FS improves for all schemes only for Gemma-7b and Llama-3. For Mixtral-8x7B Instruct, Mistral-7B-Instruct, and Gemma-7b-it the overall \textit{accuracy} increases only for a subset of selected schemes, while for others it drops significantly.
This inconsistency is demonstrated by the fact that Mixtral-8x7B Instruct achieves worse accuracy in FS compared to ZS on both tasks. 
Overall, the results on the FS setting highlight distinct strengths and weaknesses that are highly dependent on the prompt strategy, the pre-training setup, and the model's family.
At the same time, improvements observed in FS for some models might indicate a potential for enhanced performance when contextual information is available, although this gain is not universal across the models.

\paragraph{Significant variability in model accuracy is observed across different syllogistic schemes.} In general, we found that LLMs exhibit variable performances depending on the specific syllogistic schema used for evaluation. The results indicate that LLMs perform better on \textit{generalized modus ponens} (i.e., three out of four models), with \textit{accuracy} levels ranging from 0.71 to 0.98 in the FS setting in Task 1 (Fig. \ref{fig:task1_byvariant}, Table \ref{tab:task1_byscheme} in Appendix \ref{sec:appendix_tables}) and with \textit{reasoning accuracy} scores ranging from 0.56 to 0.74 in Task 2 (Table \ref{tab:task2_ra_byscheme} in Appendix \ref{sec:appendix_tables}). On the other side, In the ZS setting, the Gemma-7B-it model achieved higher performance on \textit{generalized modus tollens} (accuracy = 0.78 in Task 1 and RA = 0.74 in Task 2), while in the FS setting, the highest performances were registered on \textit{generalized dilemma} in Task 1 (accuracy = 0.85) and \textit{generalized modus ponens} in Task 2 (RA = 0.74).

\subsection{Robustness to Lexical Variations}

 We performed a robustness analysis adopting surface-form variations applied to the syllogistic schemes. This is done by rephrasing the natural language syllogistic arguments via negation, complex predicates, and De Morgan's laws, keeping the underlying logical relation between premises and conclusion unaltered (see Fig. \ref{fig:corpus_creation} in Appendix \ref{sec:appendix_pipeline}). The results of the analysis are reported in Figure \ref{fig:task1_byvariant}.

This intervention reveals that LLMs are highly sensitive to surface forms and lexical variations, showing a significant variability across different schemes (Fig. \ref{fig:task1_byvariant}). 
At the same time, we found that in the FS setting, model responses demonstrated overall greater consistency, with lower fluctuations across variants.

In Task 1, \textit{disjunctive syllogism} was generally identified as the most challenging scheme in the ZS setting for all four models, with none exceeding an accuracy of 0.35, significantly worse than random guessing. While FS slightly improved accuracy, disjunctive syllogism remained the most challenging scheme.
Overall, only Gemma-7B in the FS setting effectively handled all syllogistic schemes and their variants in Task 1, demonstrating a higher level of consistency. The other models exhibited significant variability depending on the specific surface form, which highlights a shared inability to systematically abstract the underlying reasoning rules required to derive valid conclusions.

\subsection{Impact of Distractors and Factuality}

Finally, we perform an analysis introducing an increasing number of distractors (i.e., irrelevant premises) in the syllogistic arguments and replacing existing genes in Reactome with synthetic gene names to generate arguments that are independent of medical content and assess the impact of factuality on reasoning (see Appendix \ref{sec:appendix_results_distractors}, \ref{sec:appendix_results_dummy}). 

We found that LLMs show varied sensitivity to distractors, with models like Gemma-7b exhibiting a significant decline in reasoning accuracy as the number of distractors increases, while Mistral-7B Instruct shows improvements in some cases. The impact is scheme-dependent and reveals that more complex syllogisms are particularly affected by increasing distracting information (Figs. \ref{fig:task1_bydistractors_zeroshot}-\ref{fig:task2_bydistractors_icl} in Appendix \ref{sec:appendix_figures}). 

Furthermore, contrary to previous work showing that LLMs exhibit higher performance on syllogisms that are in line with commonsense knowledge (\cite{eisape-etal-2024-systematic,kim2024mechanistic}), we found stable performances when intervening on the factual correctness of the biomedical arguments. These results indicate that (1) logical structure and contextual information have a greater impact on output generation than biomedical knowledge in both ZS and FS settings and (2) existing models might have limited exposition to human genome pathways information during pre-training, being unaffected by the substitution of real gene names with synthetic ones (Fig. \ref{fig:combined_paired_plot_comparison} in Appendix \ref{sec:appendix_results_dummy}).


\section{Related Work}

Syllogistic reasoning, which involves deriving valid conclusions from given premises based on formal logical structures, has long been a central focus in the study of Natural Language Inference (NLI) (\cite{bertolazzi2024systematicanalysislargelanguage,kim2024mechanistic}). 

Early transformer models like BERT (\cite{devlin-etal-2019-bert}) and RoBERTa (\cite{Liu2019RoBERTaAR}) were trained on general NLI datasets, such as SNLI (\cite{bowman-etal-2015-large}) and MNLI (\cite{williams-etal-2018-broad}), to address reasoning tasks. However, these models were limited in their ability to generalize abstract reasoning patterns, particularly in more specialized domains. Recent work has explored whether larger, decoder-based language models are capable of capturing syllogistic reasoning without task-specific fine-tuning, as reasoning itself is content-independent (\cite{bertolazzi2024systematicanalysislargelanguage}).

Recent research, such as Liu et al. (\cite{liu2023gloreevaluatinglogicalreasoning}), has introduced benchmarks like GLoRE to test logical reasoning in LLMs, utilizing techniques such as Chain-of-Thought (CoT) prompting (\cite{wei2023chainofthoughtpromptingelicitsreasoning}) and in-context learning (ICL) (\cite{huang-chang-2023-towards}). These methods have been shown to improve reasoning performance, but issues remain, as models often rely on superficial patterns instead of deep logical comprehension. Eisape et al. (\cite{eisape-etal-2024-systematic}) further demonstrated that even advanced models exhibit biases when handling syllogisms, particularly when faced with invalid syllogisms, often failing to generate "nothing follows" conclusions.

While the general-domain performance of LLMs in syllogistic reasoning has been explored (\cite{eisape-etal-2024-systematic, dasgupta2024languagemodelshumanlikecontent}), research on syllogistic reasoning in the biomedical domain remains scarce. This paper addresses this gap by focusing specifically on biomedical applications, where logical reasoning over specialized knowledge is critical. We introduce a large-scale biomedical syllogism dataset, SylloBio-NLI, and provide a comprehensive evaluation of the syllogistic reasoning capabilities of state-of-the-art LLMs within this domain.

\section{Conclusions}

In this work, we proposed a novel methodological framework, SylloBio-NLI, designed to evaluate the syllogistic reasoning capabilities of state-of-the-art LLMs within the biomedical domain. Through comprehensive analysis across 28 syllogistic schemes, we assessed the performance of eight different models under varying conditions, including zero-shot and few-shot settings.

Our results show that both techniques exhibit high sensitivity to superficial lexical variations, highlighting a dependency between reliability, models' architecture, and pre-training regime. Overall, our evaluation indicates that, while few-shot strategies have the potential to elicit syllogistic reasoning in LLMs, existing models are still far from achieving the robustness and consistency required for safe biomedical NLI applications in specialised domains.

\section{Limitations}

A key challenge for scaling our approach to different domains is its dependency on high-quality external ontologies and knowledge bases. This factor limits the scope of our analyses across biomedical domains. More efficient methods for populating the natural language syllogistic arguments could be investigated in future work, involving automated NLP methods, such as those used in RepoDB (\cite{brown2017standard}), MSI, Hetionet (\cite{himmelstein2017systematic}), DrugMechDB (\cite{gonzalez2023drugmechdb}), and INDRA (\cite{gyori2017word, bachman2023automated}), or synthetic data generation methods coupled with efficient quality checks. However, these approaches still face challenges in balancing precision and generalization, particularly for complex reasoning tasks in biomedicine. Further improvements are necessary to develop scalable resources and more adaptable NLP techniques for real-world applications.

\section*{Acknowledgments}

This work was funded by the European Union’s Horizon 2020 research and innovation programme under grant agreement No 965397 through the cancer core Europe DART project, by the Swiss National Science Foundation (SNSF) project ``NeuMath: Neural Discourse Inference over Mathematical Texts'' (200021\_204617), by the CRUK National Biomarker Centre, and supported by the Manchester Experimental Cancer Medicine Centre and the NIHR Manchester Biomedical Research Centre.

\bibliographystyle{unsrtnat}
\bibliography{custom}






\appendix

\section{Formalization of the SylloBio-NLI Resource Generation Process}
\label{sec:appendix_formalization_resource}

This appendix formalises the generation process of the syllogistic inference patterns.

We start by defining the mains constructs (formal and linguistic artefacts and functions) of the underlying framework:

\begin{enumerate}
    \item \textbf{Syllogistic Scheme (\( S \))}: A logical inference pattern consisting of premises and a conclusion, \( S = \{ P_1, P_2, \dots, P_n, C \} \), where \( P_i \) is premise \( i \) and \( C \) is the conclusion.
    
    \item \textbf{Formal Argument Scheme (\( \sigma \))}: Representation of a syllogistic scheme in first-order logic (FOL), \( \sigma(S) = \{ \phi_1, \phi_2, \dots, \phi_n, \psi \} \), where \( \phi_i \) corresponds to \( P_i \) and \( \psi \) corresponds to \( C \).
    
    \item \textbf{Natural Language Template (\( \tau \))}: A natural language schema mapping each formula in \( \sigma(S) \) to a sentence template, \( \tau(\sigma(S)) = \{ \tau_1, \tau_2, \dots, \tau_n, \sigma \} \), where \( \tau_i \) is the sentence template for \( \phi_i \) and \( \sigma \) is the sentence template for \( \psi \).
    
    \item \textbf{Ontology (\( O \))}: A domain-specific knowledge base containing entities \( E \) and predicates \( \Pi \), \( O = \{ E, \Pi \} \), where \( E = \{ e_1, e_2, \dots, e_k \} \) and \( \Pi = \{ \pi_1, \pi_2, \dots, \pi_l \} \).
    
    \item \textbf{Instantiation Function (\( I \))}: A function that replaces placeholders in \( \tau \) with entities and predicates from \( O \), \( I: \tau(\sigma(S)) \times O \rightarrow \text{NL} \), where \( \text{NL} \) is the set of natural language sentences.
    
    \item \textbf{Expert Mapping Function (\( \mu_{\text{Expert}} \))}: A function provided by a domain expert to map placeholders to appropriate ontology terms, \( \mu_{\text{Expert}}: \text{Placeholders} \rightarrow E \cup \Pi \).
    
    \item \textbf{Knowledge Base (\( \text{KB} \))}: A collection of instantiated syllogistic arguments, \( \text{KB} = \{ A_1, A_2, \dots, A_m \} \), where \( A_i = \{ P_1', P_2', \dots, P_n', C' \} \) and \( P_i', C' \) are instantiated natural language sentences.
\end{enumerate}

\subsection{Process Formalisation}

The process formalisation defines a systematic process for generating domain-specific syllogistic arguments by:

\begin{enumerate}
    \item \textbf{Defining} formal representations of syllogistic schemes in first-order logic.
    \item \textbf{Generating} natural language templates from these formal representations.
    \item \textbf{Mapping} placeholders to domain-specific entities and predicates using an ontology and expert knowledge.
    \item \textbf{Instantiating} the templates to produce logically valid and semantically sound arguments.
    \item \textbf{Constructing} a knowledge base for evaluating NLI models.
\end{enumerate}

This ensures that the generated arguments are both logically valid and contextually relevant to the biomedical domain.

\noindent\textbf{Input:} A set of syllogistic schemes: \( \mathcal{S} = \{ S_1, S_2, \dots, S_m \} \), an ontology: \( O = \{ E, \Pi \} \), an expert mapping function: \( \mu_{\text{Expert}} \). \\
\noindent\textbf{Output:} A knowledge base of instantiated arguments: \( \text{KB} \).\\

\noindent\textbf{Step 1: Formal Argument Scheme Selection:} For each syllogistic scheme \( S_i \in \mathcal{S} \), define its formal argument scheme in first-order logic:\\
\[
\forall S_i \in \mathcal{S}, \quad \sigma(S_i) = \{ \phi_1^i, \phi_2^i, \dots, \phi_n^i, \psi^i \}.
\]

\noindent\textbf{Step 2: Natural Language Template Generation:} Transform each formula in \( \sigma(S_i) \) into a natural language template:\\
\[
\forall \phi_j^i \in \sigma(S_i), \quad \tau_j^i = \tau(\phi_j^i),
\]
\[
\sigma^i = \tau(\psi^i).
\]

\noindent\textbf{Step 3: Ontology Mapping and Instantiation:} Apply the expert mapping function to select appropriate entities and predicates from the ontology: \\

\[
\forall \text{Placeholder } p \in \{\textbf{F}, \textbf{G}, \textbf{a}\}, \quad \mu_{\text{Expert}}(p) \rightarrow E \cup \Pi.
\]

\noindent Instantiate the templates:  \\
\[
P_j' = I(\tau_j^i, \mu_{\text{Expert}}),
\]
\[
C' = I(\sigma^i, \mu_{\text{Expert}}).
\]

\noindent under the following constraints:  \\
\begin{itemize}
    \item \textbf{Logical Validity}: The instantiated arguments must preserve the logical structure of \( \sigma(S_i) \).
    \item \textbf{Domain Soundness}: The selected entities and predicates must be semantically coherent within the targeted subdomain.
\end{itemize}

\noindent These constraints can be further formalised as: \\

\noindent\textbf{Logical Validity Constraint:} The instantiated argument \( A_i \) must be logically valid:\\
\[
\{ \phi_1', \phi_2', \dots, \phi_n' \} \models \psi',
\]
\noindent where \( \phi_j' \) corresponds to the logical form of \( P_j' \).\\

\noindent\textbf{Domain Soundness Constraint:} The entities and predicates used must be semantically valid within the domain: \\
\[
\forall e \in E', \pi \in \Pi', \quad \text{DomainValid}(e, \pi) = \text{True},
\]
\noindent where \( E' \subseteq E \) and \( \Pi' \subseteq \Pi \) are entities and predicates used in \( A_i \).\\

\noindent\textbf{Verification of Logical Validity:} Ensure that the instantiated premises logically entail the conclusion:\\
\[
\{ \phi_1', \phi_2', \dots, \phi_n' \} \models \psi',
\]
\noindent using logical inference rules.\\

\noindent\textbf{Verification of Domain Soundness:} Confirm that:\\

\begin{itemize}
    \item All entities and predicates are correctly used.
    \item There are no semantic contradictions.
\end{itemize}

\noindent\textbf{Step 4: Knowledge Base Construction:} Aggregate all instantiated arguments into the knowledge base: \\

\[
\text{KB} = \bigcup_{i=1}^m \{ A_i \},
\]
\noindent where:\\
\[
A_i = \{ P_1', P_2', \dots, P_n', C' \}.
\]

\noindent This is summarised with the following algorithmic outline:\\

\begin{enumerate}
    \item Initialize \( \text{KB} \leftarrow \emptyset \).
    \item For each \( S_i \in \mathcal{S} \):
    \begin{enumerate}
        \item Create \( \sigma(S_i) = \{ \phi_1^i, \phi_2^i, \dots, \phi_n^i, \psi^i \} \).
        \item Generate \( \tau(S_i) = \{ \tau_1^i, \tau_2^i, \dots, \tau_n^i, \sigma^i \} \), where \( \tau_j^i = \tau(\phi_j^i) \).
        \item Obtain mappings \( \mu_{\text{Expert}} \) for placeholders in \( \tau_j^i \) and \( \sigma^i \).
        \item Instantiate:
        \[
        P_j' = I(\tau_j^i, \mu_{\text{Expert}}),
        \]
        \[
        C' = I(\sigma^i, \mu_{\text{Expert}}).
        \]
        \item Form argument \( A_i = \{ P_1', P_2', \dots, P_n', C' \} \).
        \item Add \( A_i \) to \( \text{KB} \):
        \[
        \text{KB} \leftarrow \text{KB} \cup \{ A_i \}.
        \]
    \end{enumerate}
    \item Return \( \text{KB} \).
\end{enumerate}

\section{Domain-specific pipeline for creating NL instances}
\label{sec:appendix_pipeline}


\begin{figure*}[tb]
\centering
\includegraphics[width= .9\textwidth]{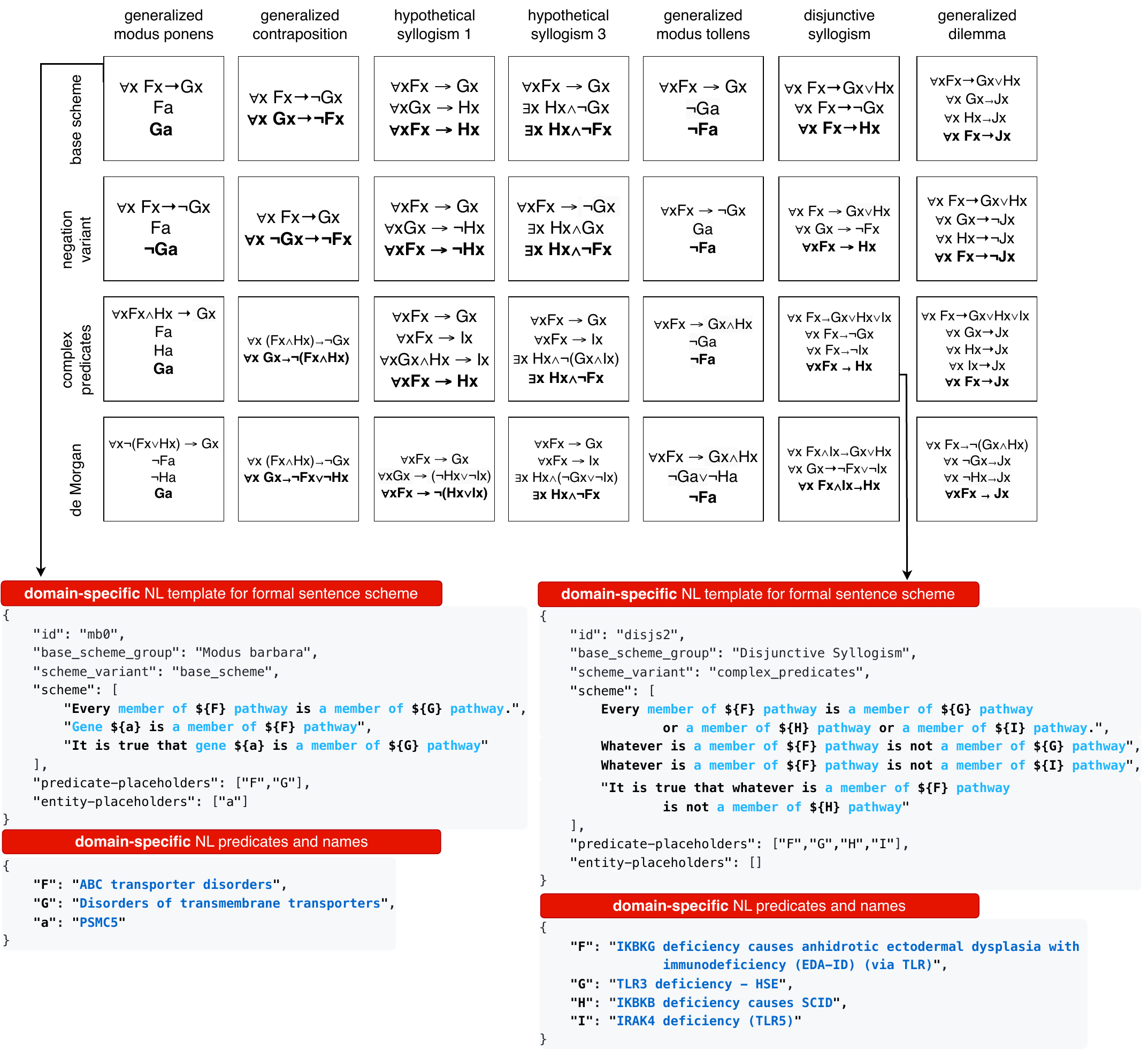}
\caption{Syllogistic argument schemes used to create a biologically factual argument corpus with domain-specific examples for \textit{generalized modus ponens base scheme} and \textit{disjunctive syllogism complex predicates scheme}. For each syllogistic scheme, a formal argument scheme (consisted of premises and conclusion (bold)) was provided.}
\label{fig:corpus_creation}
\end{figure*}

Figure \ref{fig:corpus_creation}: Syllogistic argument schemes used to create a biologically factual argument corpus with domain-specific examples for \textit{generalized modus ponens base scheme} and \textit{disjunctive syllogism complex predicates scheme}. For each syllogistic scheme, a formal argument scheme (consisted of premises and conclusion (bold)) was provided.

\section{Accessing LLMs}
\label{sec:appendix_llms}

To access these LLMs, we use the Mistral AI (mistral), and the open-source weights of the remaining models, available at the \textit{HuggingFace Hub}\footnote{\url{https://huggingface.co}} repositories: 
\begin{itemize}
  \item mistralai/Mistral-7B-v0.1
  \item mistralai/Mistral-7B-Instruct-v0.2
  \item mistralai/Mixtral-8x7B-Instruct-v0.1
  \item google/gemma-7b
  \item google/gemma-7b-it
  \item meta-llama/Meta-Llama-3-8B
  \item meta-llama/Meta-Llama-3-8B-Instruct
  \item BioMistral/BioMistral-7B
\end{itemize}

The pretrained LLM weights are used through the \textit{transformers}\footnote{\url{https://huggingface.co/docs/transformers/}} python library. All models were loaded with standard configurations and their respective default tokenizers, using the \textit{AutoModelForCausalLM} and \textit{AutoTokenizer} classes. Additionally,  the models were loaded with the options \textit{device\_map=``auto"}, \textit{torch\_dtype=``auto"} \textit{attn\_implementation=``flash\_attention\_2"} and \textit{offload\_buffers=True} to make the best use of GPU resources available.

For the instruction models, the inputs were passed through each model's chat template (with $tokenizer.apply\_chat\_template$), so that they would follow the appropriate prompt format. Responses were cleaned of special symbols for evaluation.
Table \ref{tab:models} contains the relevant characteristics of all analyzed models.

\begin{table*}[]
\resizebox{.99\textwidth}{!}{%
\begin{tabular}{llcclccclc}
\hline
\textbf{Model}             & \textbf{Type}       & \textbf{Parameters} & \textbf{Training Data}                                                                                                                                         & \textbf{Architecture Enhancements}                                                 & \textbf{Open-Source}                                               & \textbf{Performance in Benchmarks}                                                                                                                                                    & \textbf{Efficiency}                                                                                        & \textbf{Use Cases}                                                                                    & \textbf{Year} \\ \hline
Mistral-7B-v0.1            & Base LLM            & 7B                  & Diverse (internet, academic, etc.)                                                                                                                             & \begin{tabular}[c]{@{}l@{}}Grouped-query attention, \\ sliding-window\end{tabular} & Yes                                                                & \begin{tabular}[c]{@{}c@{}}Strong in reasoning, \\ real-life scenarios\end{tabular}                                                                                                   & Similar, optimizations possible                                                                            & Math, code generation                                                                                 & 2023          \\
Mistral-7B-Instruct-v0.2   & Instruct LLM        & 7B                  & Instruction-tuned data                                                                                                                                         & Instruction-tuned architecture                                                     & Yes                                                                & High on MT-Bench                                                                                                                                                                      & Efficient for instruction-based tasks                                                                      & \begin{tabular}[c]{@{}l@{}}Customer service chatbots \\ and virtual assistants\end{tabular}           & 2023          \\
Mixtral-8x7B-Instruct-v0.1 & Instruct LLM        & 8x7B                & Diverse data with expert focus                                                                                                                                 & Mixture of Experts architecture                                                    & Yes                                                                & \begin{tabular}[c]{@{}c@{}}Exceptional performance in \\ Arena Elo and MMLU, \\ excellent MT-Bench score\end{tabular}                                                                 & \begin{tabular}[c]{@{}c@{}}Great at simulated dialogues \\ and general language understanding\end{tabular} & \begin{tabular}[c]{@{}l@{}}Customer service chatbots \\ and interactive storytelling\end{tabular}     & 2023          \\
Gemma-7B                   & Base LLM            & 7B                  & 6 trillion tokens (web, math, code)                                                                                                                            & \begin{tabular}[c]{@{}l@{}}Multi-query attention, RoPE, \\ GeGLU\end{tabular}      & \begin{tabular}[c]{@{}c@{}}Yes \\ (with terms of use)\end{tabular} & Strong in code generation, math                                                                                                                                                       & Lightweight, runs on various platforms                                                                     & Text generation, translation                                                                          & 2023          \\
Gemma-7B-it                & Instruct LLM        & 7B                  & 6 trillion tokens (web, math, code)                                                                                                                            & Instruction-tuned architecture                                                     & Yes                                                                & \begin{tabular}[c]{@{}c@{}}Strong in reasoning, \\ in MMLU and HellaSwag\end{tabular}                                                                                                 & Efficient for instruction-based tasks                                                                      & \begin{tabular}[c]{@{}l@{}}Text comprehension \\ and generation inference\end{tabular}                & 2023          \\
Meta-Llama-3-8B            & Base LLM            & 8B                  & Mix of publicly available online data                                                                                                                          & Standard transformer                                                               & Yes                                                                & Moderate                                                                                                                                                                              & Efficient for general tasks                                                                                & General NLP tasks                                                                                     & 2023          \\
Meta-Llama-3-8B-Instruct   & Instruct LLM        & 8B                  & Instruction-tuned data                                                                                                                                         & Instruction-tuned architecture                                                     & Yes                                                                & \begin{tabular}[c]{@{}c@{}}High for Arena Elo, \\ impressive MT Bench scores \\ for translation, exceptional \\ MMLU score, indicating \\ strong reasoning and knowledge\end{tabular} & Efficient for instruction-based tasks                                                                      & \begin{tabular}[c]{@{}l@{}}High-volume applications, \\ great for real-time interactions\end{tabular} & 2023          \\
BioMistral-7B              & Domain-Specific LLM & 7B                  & \begin{tabular}[c]{@{}c@{}}Collection of medical LLMs \\ resulting from further pre-training \\ Mistral-7B Instruct on\\ PubMed Central resources\end{tabular} & Enhanced for domain-specific tasks                                                 & Yes                                                                & \begin{tabular}[c]{@{}c@{}}State- of-the-art performance on \\ the multilingual medical \\ evaluation benchmark \\ compared to other open-source \\ 7B models\end{tabular}            & Efficient for domain-specific tasks                                                                        & Biomedical research                                                                                   & 2024          \\ \hline
\end{tabular}%
}
\caption{\label{tab:models}
    Detailed Characteristics of Selected LLMs.
  }
\end{table*}

\section{Experimental Details}
\label{sec:appendix_experimental}

The entire experimental setup was implemented as a \textit{python} code package, consisting of 5 modules:

\begin{itemize}
  \item \textit{pathways}: defines the ontological relations and operations for biological pathways, as well as the logic to retrieve and transform data from the domain ontology.
  \item \textit{logic2nl}: translates logic formulas into NL statements (premises, conclusions), using parameterised scheme templates.
  \item \textit{llm}: provides access to LLMs, and facilitates logging.
  \item \textit{experiments}: defines all test logic, parameterisation and metrics for the experiments.
  \item \textit{main}: orchestrates all the tests and aggregates results.
\end{itemize}

The LLMs were loaded and run using the HuggingFace transformers library (v4.43.3). The prompts were processed directly by the models with \textit{generate}, without sampling.

Parameters were set as follows:

\begin{itemize}
  \item \textit{Number of premises} = $2$: the number of valid premises per instance (factual argumentative text).
  \item \textit{Max distractors} = $5$: maximum number of distractors to be added per instance.
  \item \textit{Subset size} = $200$: maximum number of positive and negative instances for each scheme.
  \item \textit{Batch size} = $20$: number of instances evaluated simultaneously.
  \item \textit{ICL}: Whether the in-context learning prompt would be used or not.
  \item \textit{Model}: the LLM to be evaluated.
\end{itemize}



For both Task 1 and Task 2, each model was analyzed across all 28 syllogistic schemes, with responses evaluated from a total of 11,200 instances (28 schemes × (200 positive + 200 negative)). The exceptions were: \textit{Generalized Modus Ponens} - complex predicates (18 instances), \textit{Hypothetical Syllogism 1} - base (202 instances) and \textit{Generalized Contraposition} - negation (202 instances). This is due to the number of factually possible different instances being limited by the ground-truth data in such cases. For both tasks, this amounted to a total of 211,840 instances, with performance comparisons made between ZS and ICL settings, leading to a total of 423,680 prompts.

When examining the impact of distractors, responses were analyzed across five variants of each argument scheme, reflecting different numbers of irrelevant premises (from 1 to 5 distractors). For each scheme and model, this resulted in 1,000 responses (200 prompts per scheme × 5 variants), with the exception of the aforementioned cases. Across all schemes, this totaled 26,480 responses for one model, and 211,840 responses were analyzed for all models for each number of distractors, totaling 1,059,200 responses. Again, comparisons were made between ZS and ICL settings, doubling the previous number.

For both Task 1 and Task 2, performance was compared using actual gene entity names versus synthetic names across two selected schemes. Each scheme was tested with 400 instances (200 positive, 200 negative), resulting in a comprehensive analysis of 12,800 instances per task (200 instances × 2 schemes × 2 entity types × 8 models), providing a thorough comparison of model performance with both factual and synthetic gene names.

The experiments were run on a computer with an AMD EPYC 7413 24-Core CPU, 128GB of available RAM and 2 $\times$ NVIDIA A100-SXM4-80GB GPUs.

\section{Evaluation Metrics}
\label{sec:appendix_evaluation}
\textbf{Accuracy, F1-score, Recall, and Precision} are used to evaluate performance in Task 1, a binary classification task where the conclusion \( C \) is labeled as either True or False. These metrics compare the predicted labels \( \hat{y}_i \) against the annotated gold labels \( y_i \) (True/False).

\textbf{Reasoning Accuracy} $RA$ evaluates a model's ability in Task 2 to correctly predict the entailment label \( \hat{y}_i \) and select the appropriate subset of premises \( \hat{P}_i \) that justify this prediction. The metric is calculated as the percentage of responses where both the predicted label \( \hat{y}_i \) matches the ground truth \( y_i \) and the selected premises \( \hat{P}_i \) match the ground truth premises \( P_i \).
\[
\text{RA} = \frac{1}{N} \sum_{i=1}^{N} \mathbb{I}\left( \hat{y}_i = y_i \ \text{and} \ \hat{P}_i = P_i \right)
\]
In addition we used two supplementary metrics:
\textbf{Non-empty output} quantifies the frequency with which the model generates any textual response. This metric is crucial, as an empty output, while not providing an incorrect answer, indicates a failure to engage with the prompt.
\textbf{Irrelevant text} assesses the extent to which the model's output deviates from the expected format, such as failing to adhere to the directive to respond with a straightforward "True" or "False", or a structured statement like "The correct answer is True." This metric identifies instances where the model’s response is extraneous or non-compliant with the specified guidelines.

\textbf{Faithfulness} $F$ evaluates a model's ability to consistently adjust its predictions in response to meaningful changes in input. In this analysis, faithfulness is measured by examining $N$ pairs of prompts that differ only in the 'C' conclusion: one concludes with "it is True that..." ($y_i$) and the other with "it is False that..." ($x_i$). A pair is deemed faithful if the model changes its prediction ($f(x_i)$,$f(y_i)$) accordingly—shifting from True to False or from False to True between the two prompts.
If the model's response to either prompt in a pair is empty or contains irrelevant text that does not follow the instructions, the entire pair is considered unfaithful ($g=0$). This criterion ensures that only responses that are meaningful and adhere to the given instructions are evaluated.
\[
F = \frac{1}{N} \sum_{i=1}^{N} \mathbb{I}\left( f(x_i) \neq f(y_i) \right) \cdot \text{g}(f(x_i), f(y_i))
\]

\section{Prompting LLMs - Zero-shot prompts}
\label{sec:appendix_prompting_zs}

\subsection{TASK 1}

prompt = \textit{Suppose you are a specialist with existing knowledge about a signaling and metabolic molecules and their relations organized into biological pathways and processes. Given premises marked with the letter P and the following number and the conclusion marked with the letter C, determine whether the conclusion logically follows from these premises. If the conclusion logically follows from the premises, you need to return 'True'. If the conclusion does not follow logically from the premises, you need to return 'False'. The output should be a single word <True> or <False>.}

"\textit{P1:} " + Premise 1

"\textit{P2:} " + Premise 2

"\textit{C:}" + Conclusion
\\
\subsection{TASK 2}

prompt = \textit{Suppose you are a specialist with existing knowledge about a signaling and metabolic molecules and their relations organized into biological pathways and processes. Given premises marked with the letter P and the following number and the conclusion marked with the letter C, determine whether the conclusion logically follows from these premises. If the conclusion logically follows from the premises, you need to return 'True'. If the conclusion does not follow logically from the premises, you need to return 'False'. Specify the premises you used to determine whether the conclusion logically follows from the premises, and only these premises. The output should be a single word <True> or <False> and the numbers of the selected premises after the decimal point, like <True, P1, P2>.}

"\textit{P1:} " + Premise 1

"\textit{P2:} " + Premise 2

"\textit{C:}" + Conclusion
\\

\section{Prompting LLMs - Few-shot prompts}
\label{sec:appendix_prompting_fs}

\subsection{TASK 1}

Context: \textit{Suppose you are a specialist with existing knowledge about a signaling and metabolic molecules and their relations organized into biological pathways and processes.}
\\
Instructions: \textit{Given premises marked with the letter P and the following number and the conclusion marked with the letter C, determine whether the conclusion logically follows from these premises.}
\\
Relevance: \textit{If the conclusion logically follows from the premises, you need to return 'True'. If the conclusion does not follow logically from the premises, you need to return 'False'.}
\\
Constraint: \textit{The output should be a single word <True> or <False>.}
\\
Demonstration: \textit{}
\\

"\textit{P1:} " \textit{"Every member of Diseases of hemostasis pathway is a member of Disease pathway"}

"\textit{P2:} " \textit{"Gene GP1BB is a member of Diseases of hemostasis pathway"}

"\textit{C:}" \textit{"Gene GP1BB is a member of Disease pathway"}

The correct answer is: True
\\

"\textit{P1:} " \textit{"Every member of Infectious disease pathway is a member of Disease pathway"}

"\textit{P2:} " \textit{"Gene PKQQ is a member of Infectious disease pathway"}

"\textit{C:}" \textit{"Gene PKQQ is a member of Infectious disease pathway"}

The correct answer is: True
\\

"\textit{P1:} " \textit{"Every member of SLC transporter disorders pathway is a member of Disorders of transmembrane transporters pathway"}

"\textit{P2:} " \textit{"Gene AXZY is a member of SLC transporter disorders pathway"}

"\textit{C:}" \textit{"Gene AXZY is not a member of Disorders of transmembrane transporters pathway"}

The correct answer is: False
\\

"\textit{P1:} " \textit{"Every member of HIV Life Cycle pathway is a member of HIV Infection pathway"}

"\textit{P2:} " \textit{"Gene MLLX is a member of HIV Life Cycle pathway"}

"\textit{C:}" \textit{"Gene MLLW is a member of HIV Infection pathway"}

The correct answer is: False
\\

"\textit{P1:} " \textit{"Every member of ABC transporter disorders pathway is a member of Disorders of transmembrane transporters pathway"}

"\textit{P2:} " \textit{"Gene PSMC5 is a member of ABC transporter disorders pathway"}

"\textit{C:}" \textit{"It is true that Gene PSMC5 is a member of Disorders of transmembrane transporters pathway"}

\subsection{TASK 2}

Context: \textit{Suppose you are a specialist with existing knowledge about a signaling and metabolic molecules and their relations organized into biological pathways and processes.}
\\
Instructions: \textit{Given premises marked with the letter P and the following number and the conclusion marked with the letter C, determine whether the conclusion logically follows from these premises.}
\\
Relevance: \textit{If the conclusion logically follows from the premises, you need to return 'True'. If the conclusion does not follow logically from the premises, you need to return 'False'. Specify the premises you used to determine whether the conclusion logically follows from the premises, and only these premises.}
\\
Constraint: \textit{The output should be a single word <True> or <False> and the numbers of the selected premises after the decimal point, like <True, P1, P2>.}
\\
Demonstration: \textit{}
\\

"\textit{P1:} " \textit{"Every member of Diseases of hemostasis pathway is a member of Disease pathway"}

"\textit{P2:} " \textit{"Every member of NS1 Mediated Effects on Host Pathways pathway is a member of Influenza Infection pathway"}

"\textit{P3:} " \textit{"Gene AABC is a member of Diseases of hemostasis pathway"}

"\textit{C:}" \textit{"Gene AABC is a member of Disease pathway"}

The correct answer is: True, P1, P3
\\

"\textit{P1:} " \textit{"Every member of SARS-CoV Infections pathway is a member of Viral Infection Pathways pathway"}

"\textit{P2:} " \textit{"Every member of Infectious disease pathway is a member of Disease pathway"}

"\textit{P3:} " \textit{"Gene PKQQ is a member of Infectious disease pathway"}

"\textit{C:}" \textit{"Gene PKQQ is a member of Infectious disease pathway"}

The correct answer is: True, P2, P3
\\

"\textit{P1:} " \textit{"Every member of SLC transporter disorders pathway is a member of Disorders of transmembrane transporters pathway"}

"\textit{P2:} " \textit{"Gene AXZY is a member of SLC transporter disorders pathway"}

"\textit{C:}" \textit{"Gene AXZY is not a member of Disorders of transmembrane transporters pathway"}

The correct answer is: False
\\

"\textit{P1:} " \textit{"Every member of HIV Life Cycle pathway is a member of HIV Infection pathway"}

"\textit{P2:} " \textit{"Gene MLLX is a member of HIV Life Cycle pathway"}

"\textit{C:}" \textit{"Gene MLLW is a member of HIV Infection pathway"}

The correct answer is: False
\\

"\textit{P1:} " \textit{"Every member of ABC transporter disorders pathway is a member of Disorders of transmembrane transporters pathway"}

"\textit{P2:} " \textit{"Gene PSMC5 is a member of ABC transporter disorders pathway"}

"\textit{C:}" \textit{"It is true that Gene PSMC5 is a member of Disorders of transmembrane transporters pathway"}

\section{Results: Misaligned Instruction-Response}
\label{sec:appendix_results_responses}
\begin{figure*}[t]
    \centering
    \includegraphics[width=0.22\linewidth]{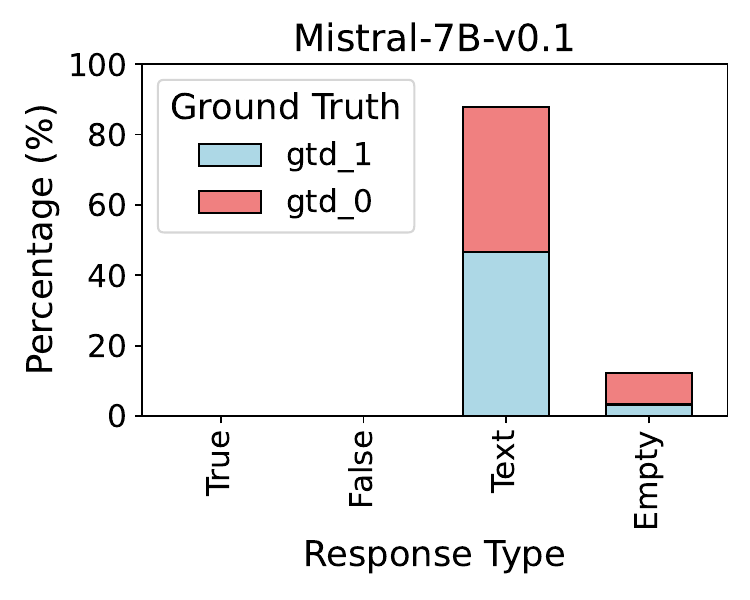} \hfill
    \includegraphics[width=0.22\linewidth]{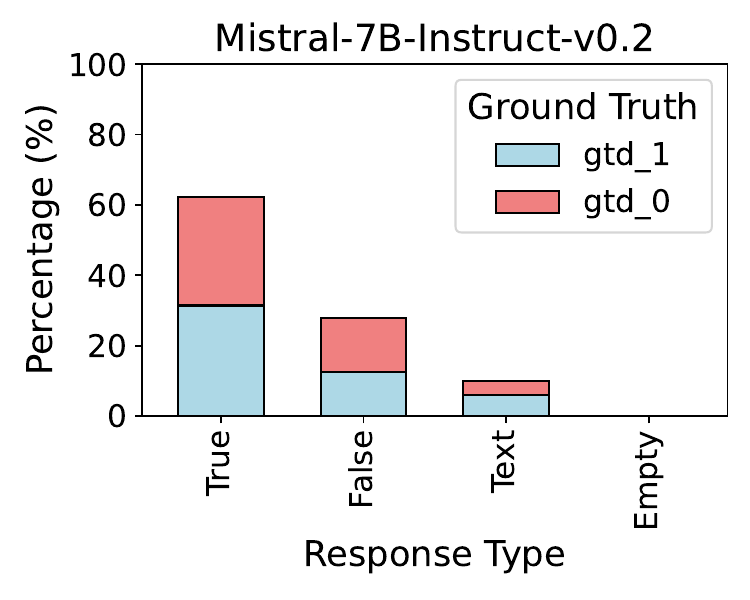} \hfill
    \includegraphics[width=0.22\linewidth]{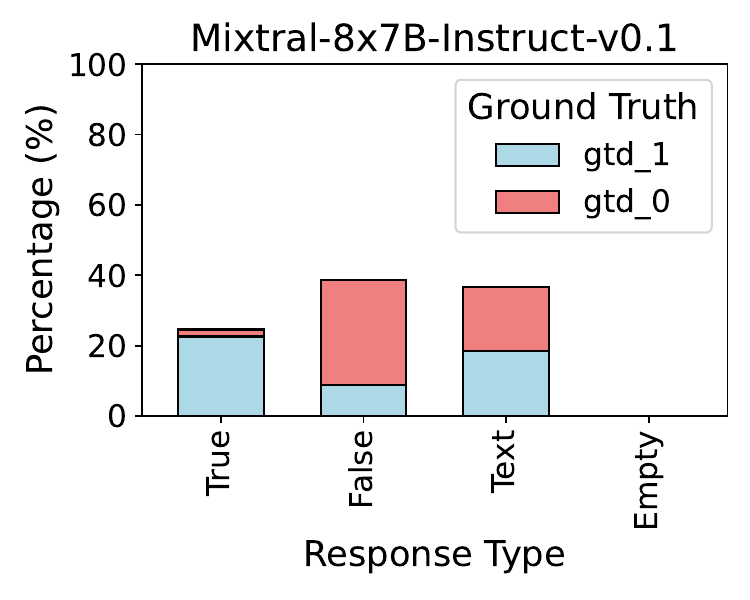} \hfill
    \includegraphics[width=0.22\linewidth]{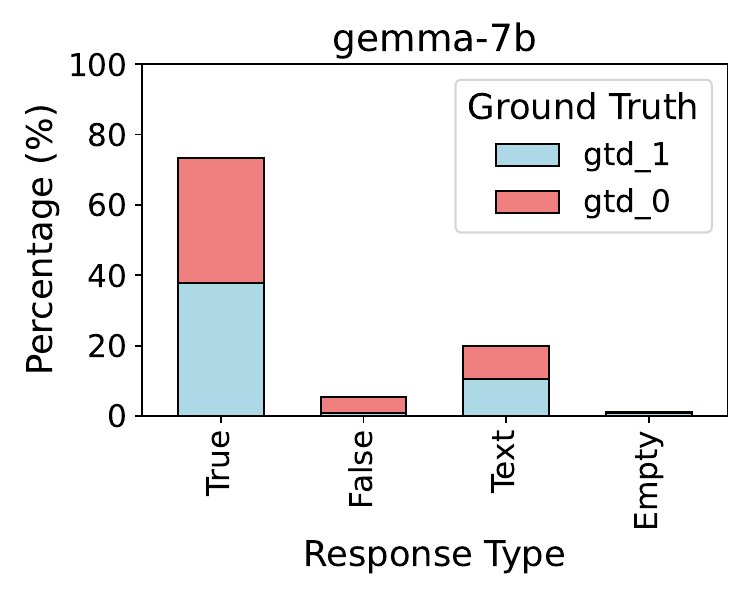} \\
    
    \includegraphics[width=0.22\linewidth]{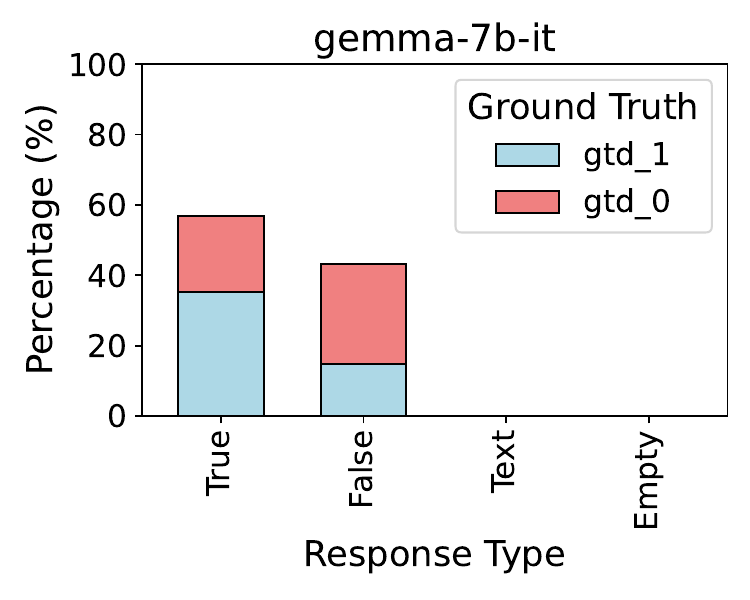} \hfill
    \includegraphics[width=0.22\linewidth]{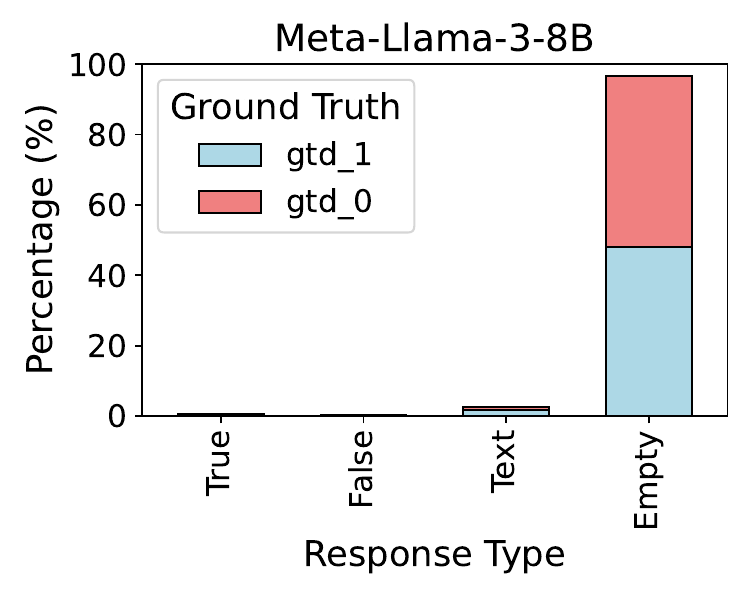} \hfill
    \includegraphics[width=0.22\linewidth]{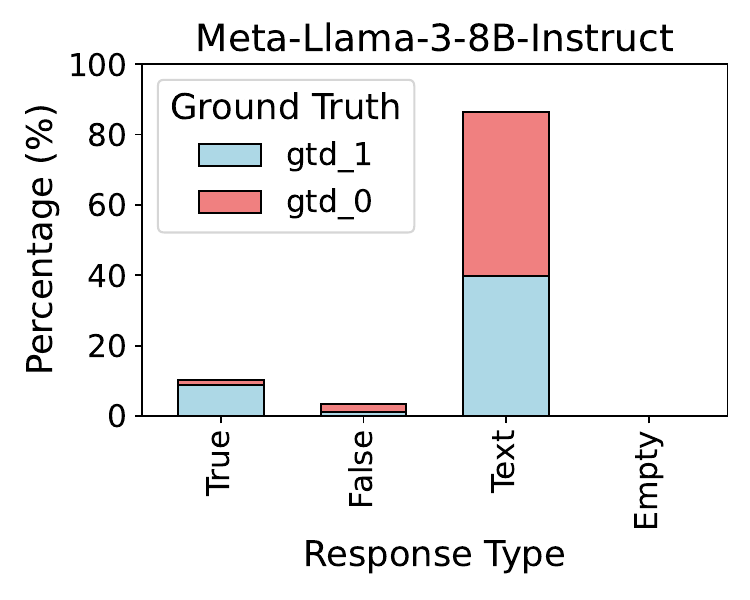} \hfill
    \includegraphics[width=0.22\linewidth]{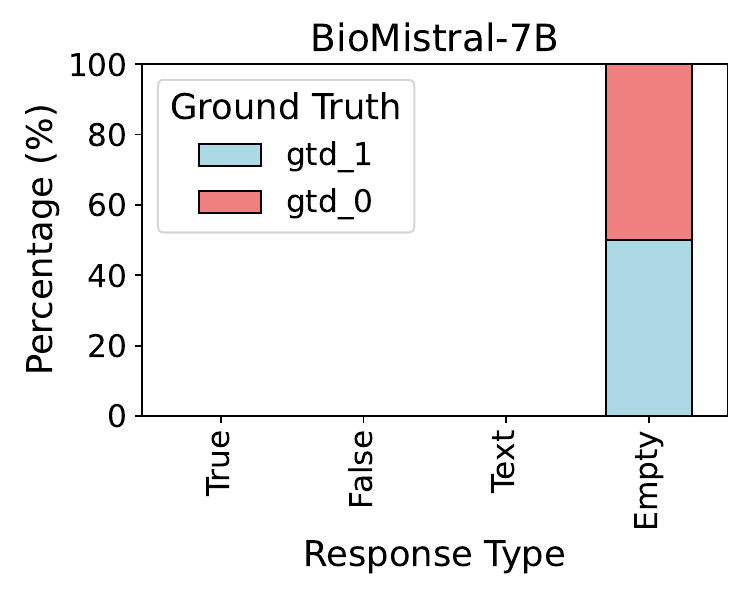}
    
    \caption{Percentage distribution of model response types under zero-shot settings for prompts with no distractors for the set of biologically factual argumentative texts.}
    \label{fig:eight_plots_zeroshot}
\end{figure*}

\begin{figure*}[t]
    \centering
    \includegraphics[width=0.22\linewidth]{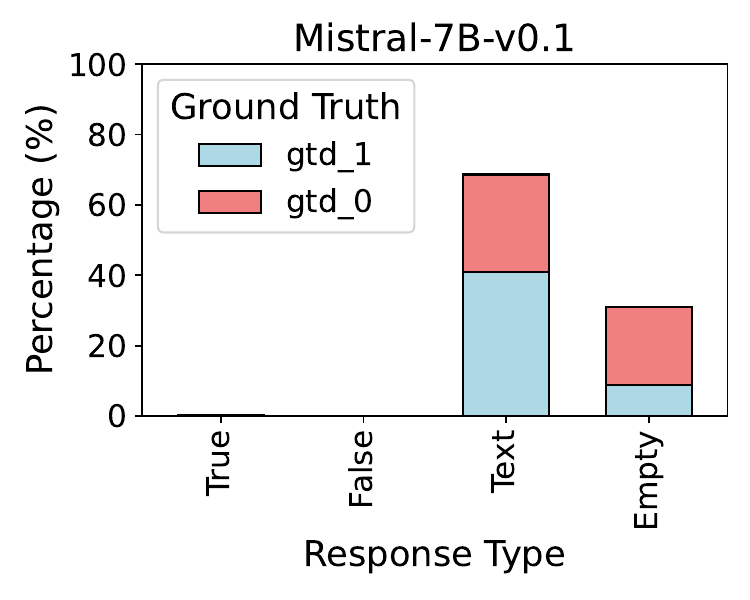} \hfill
    \includegraphics[width=0.22\linewidth]{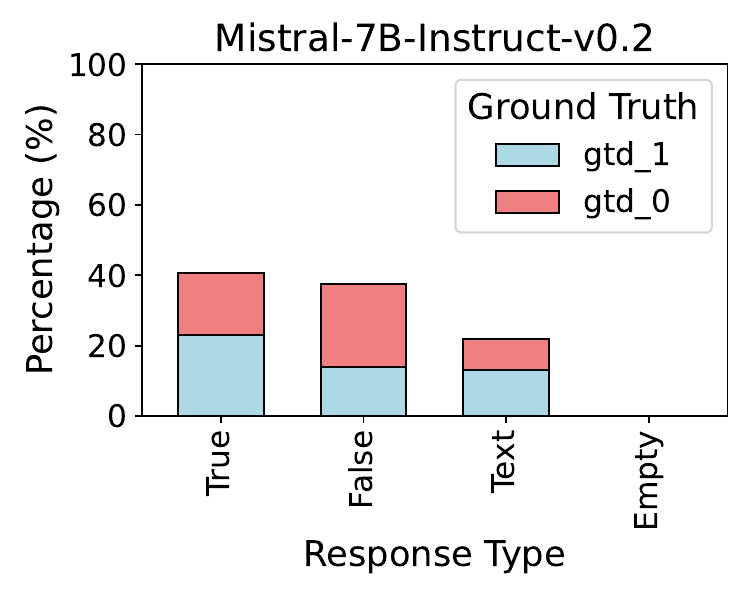} \hfill
    \includegraphics[width=0.22\linewidth]{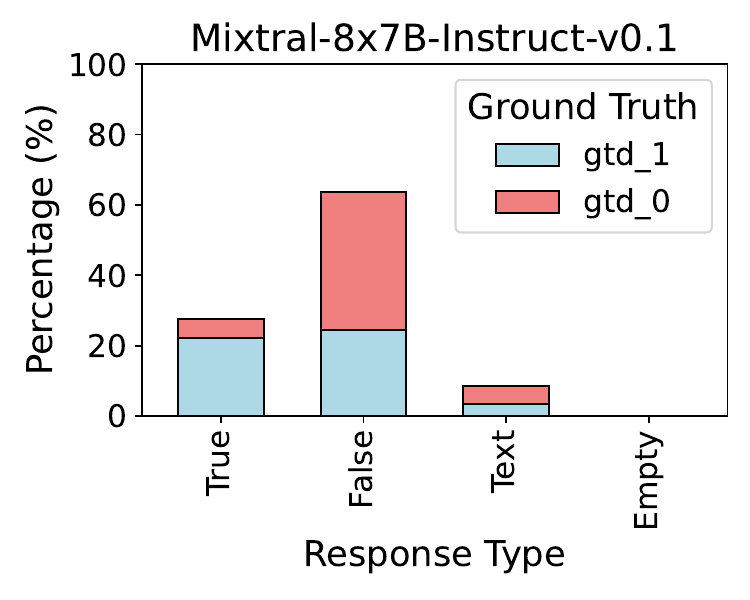} \hfill
    \includegraphics[width=0.22\linewidth]{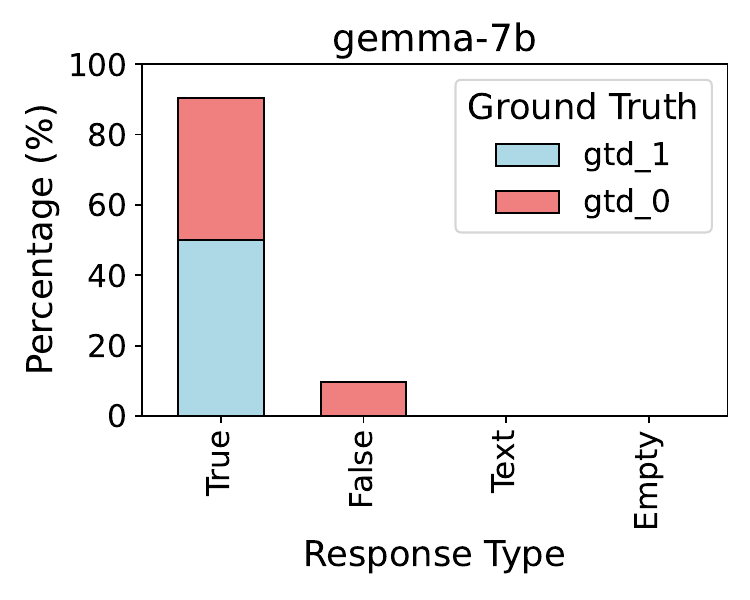} \\
    
    \includegraphics[width=0.22\linewidth]{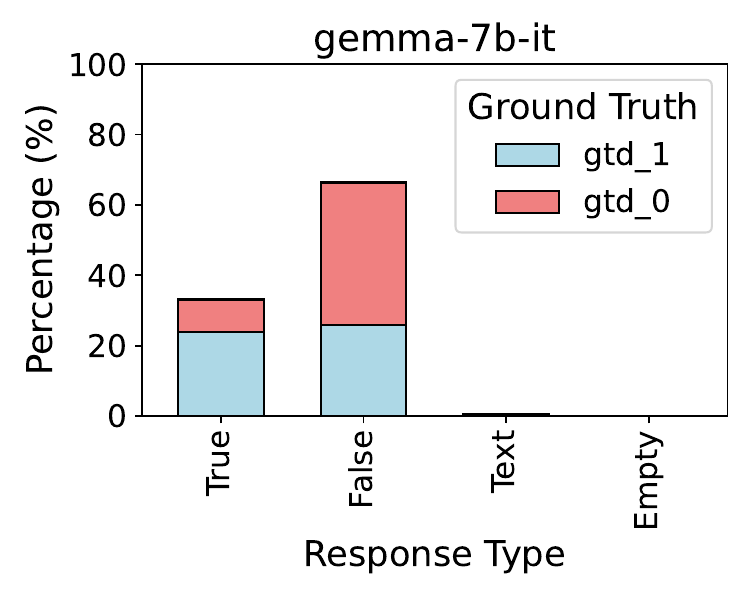} \hfill
    \includegraphics[width=0.22\linewidth]{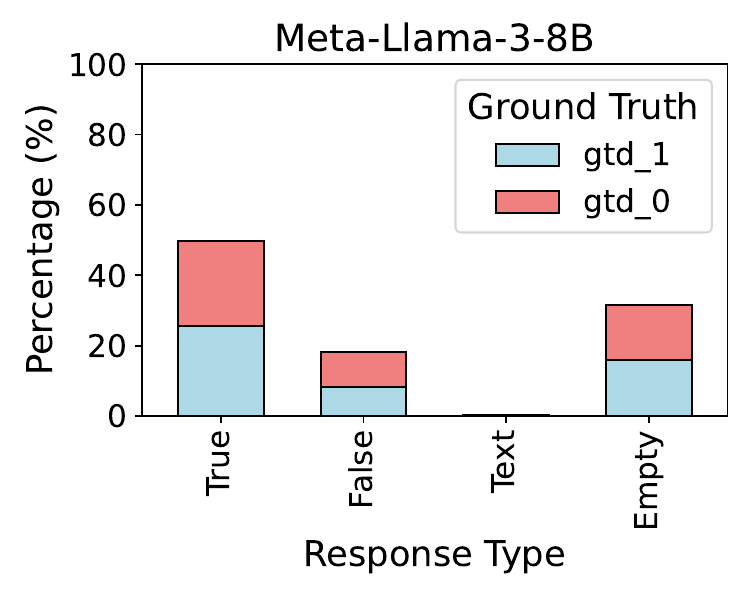} \hfill
    \includegraphics[width=0.22\linewidth]{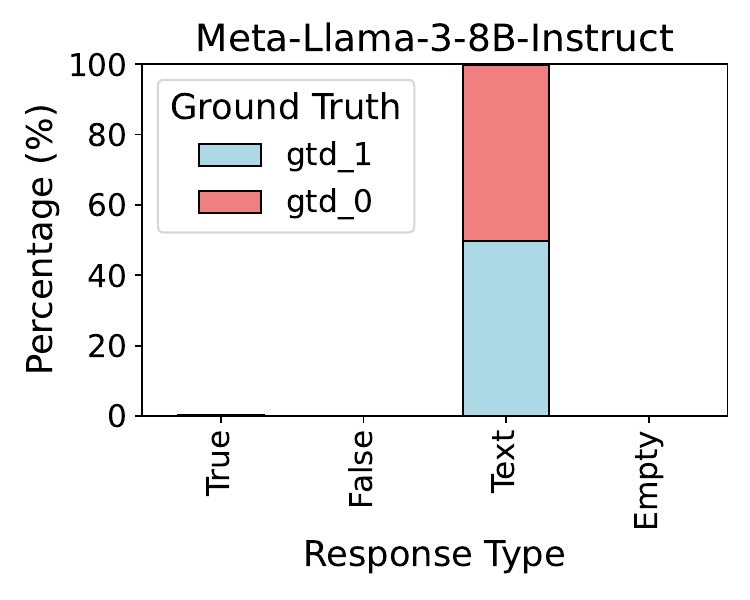} \hfill
    \includegraphics[width=0.22\linewidth]{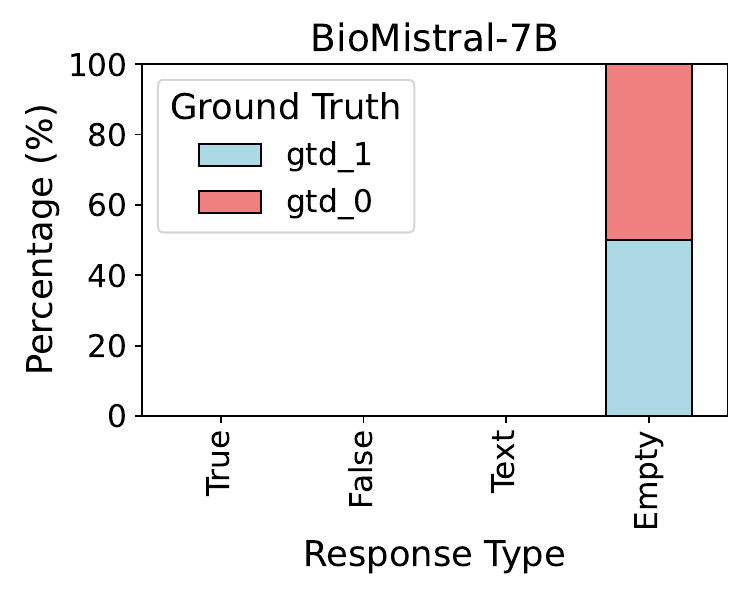}
    
    \caption{Percentage distribution of model response types under few-shot settings for prompts with no distractors for the set of biologically factual argumentative texts.}
    \label{fig:eight_plots_icl}
\end{figure*}

\begin{figure*}[t]
    \centering
    \includegraphics[width=0.22\linewidth]{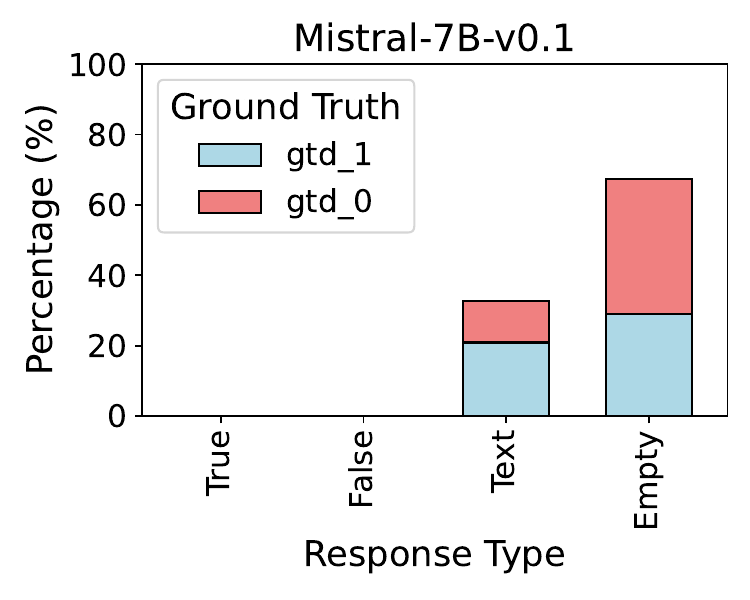} \hfill
    \includegraphics[width=0.22\linewidth]{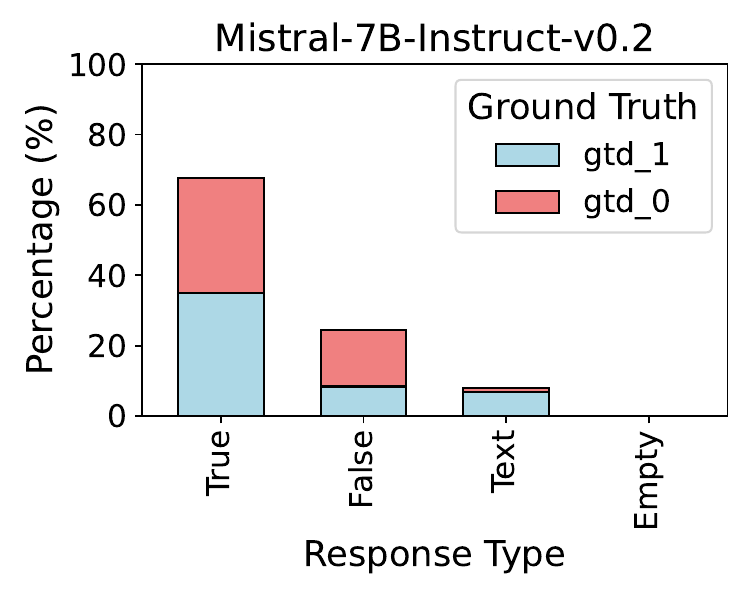} \hfill
    \includegraphics[width=0.22\linewidth]{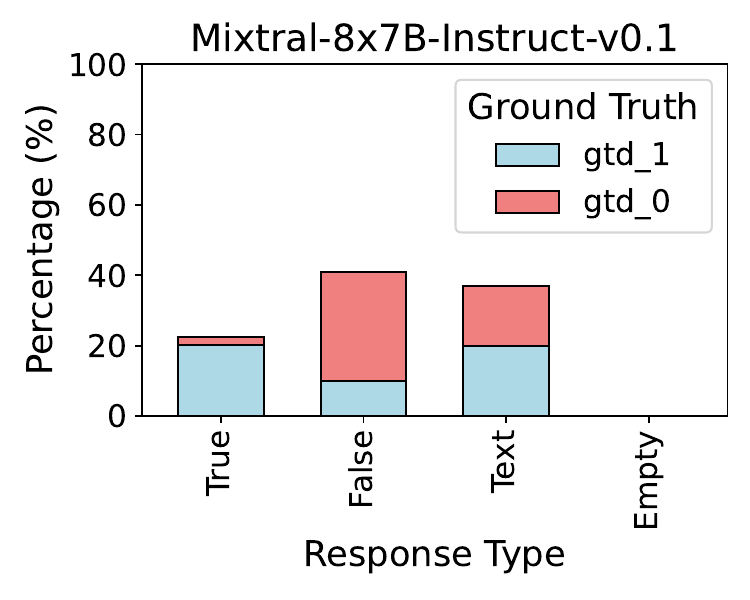} \hfill
    \includegraphics[width=0.22\linewidth]{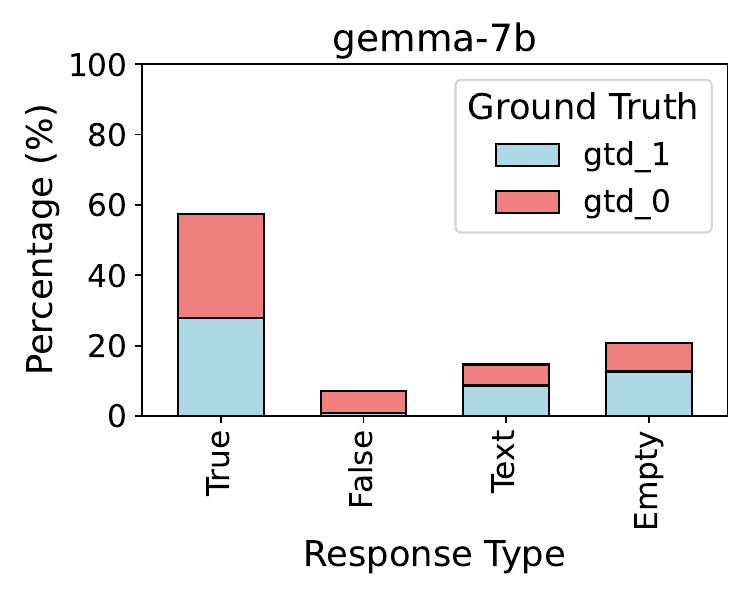} \\
    
    \includegraphics[width=0.22\linewidth]{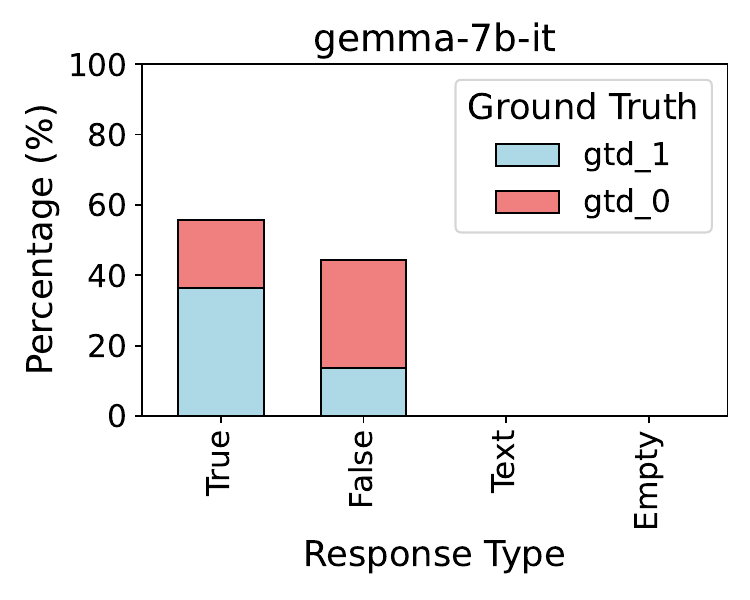} \hfill
    \includegraphics[width=0.22\linewidth]{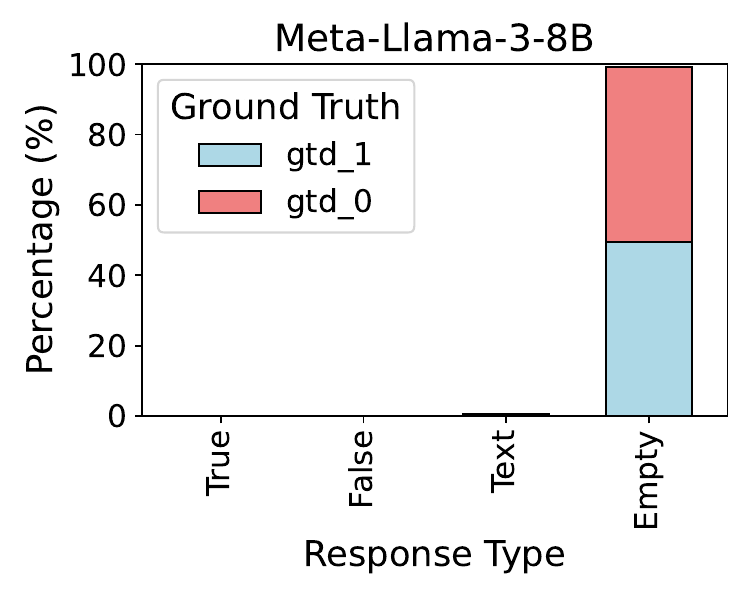} \hfill
    \includegraphics[width=0.22\linewidth]{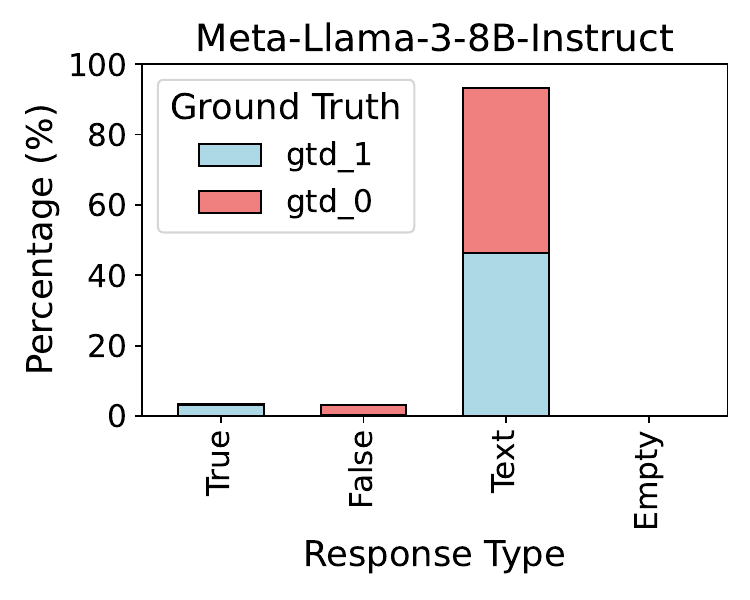} \hfill
    \includegraphics[width=0.22\linewidth]{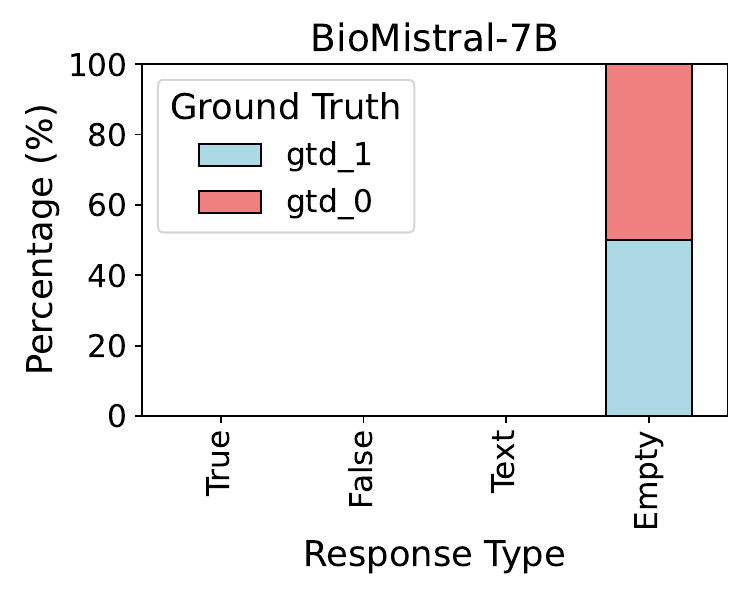}
    
    \caption{Percentage distribution of model response types under zero-shot settings for prompts with all distractors for the set of biologically factual argumentative texts.}
    \label{fig:eight_plots_all_zeroshot}
\end{figure*}

\begin{figure*}[t]
    \centering
    \includegraphics[width=0.22\linewidth]{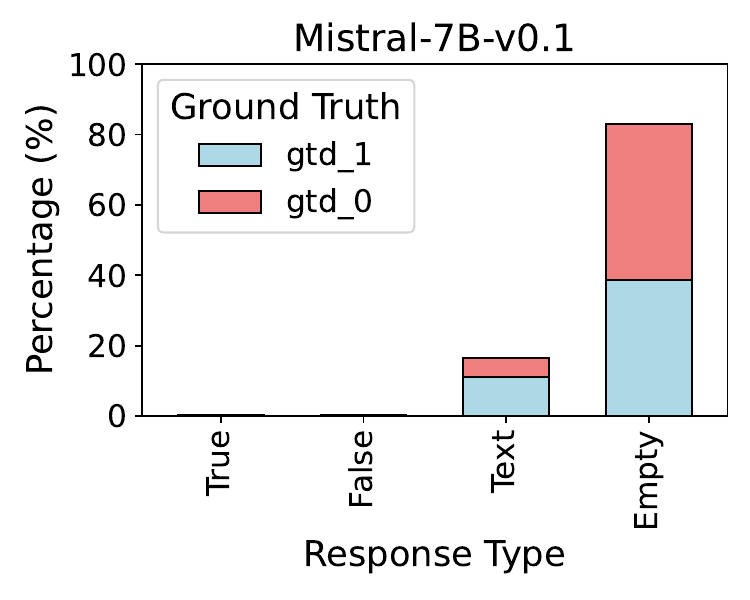} \hfill
    \includegraphics[width=0.22\linewidth]{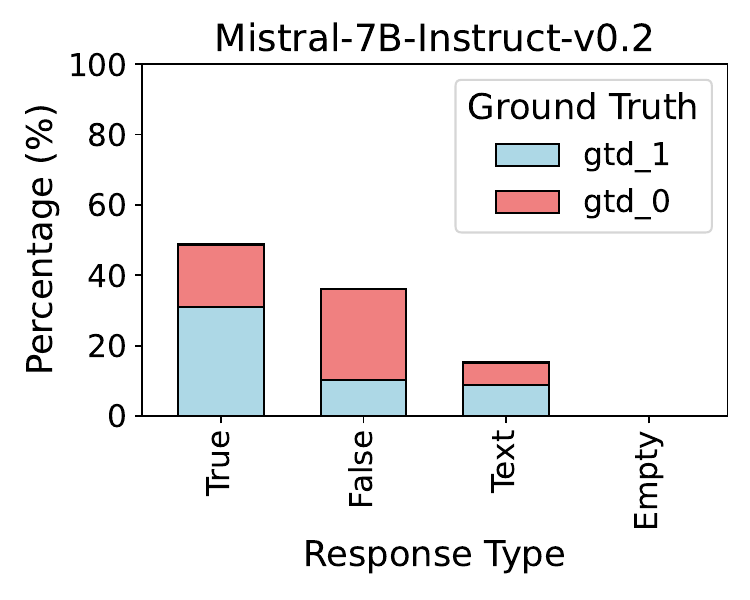} \hfill
    \includegraphics[width=0.22\linewidth]{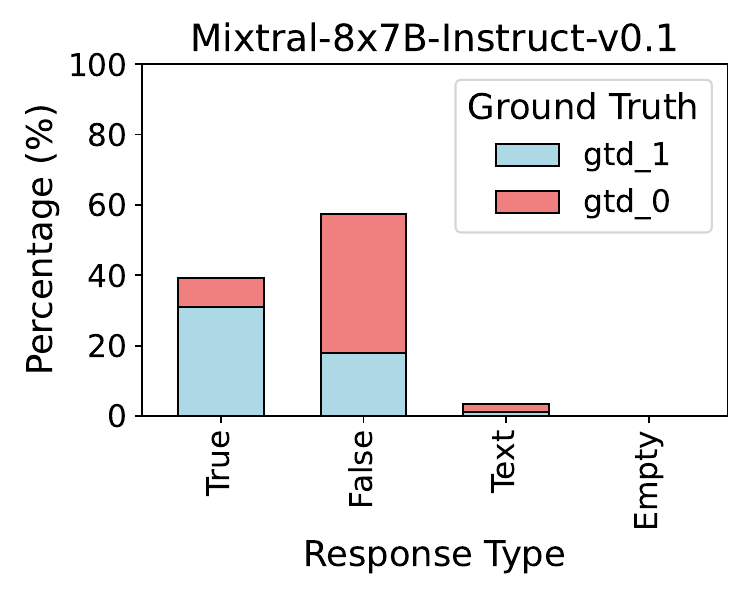} \hfill
    \includegraphics[width=0.22\linewidth]{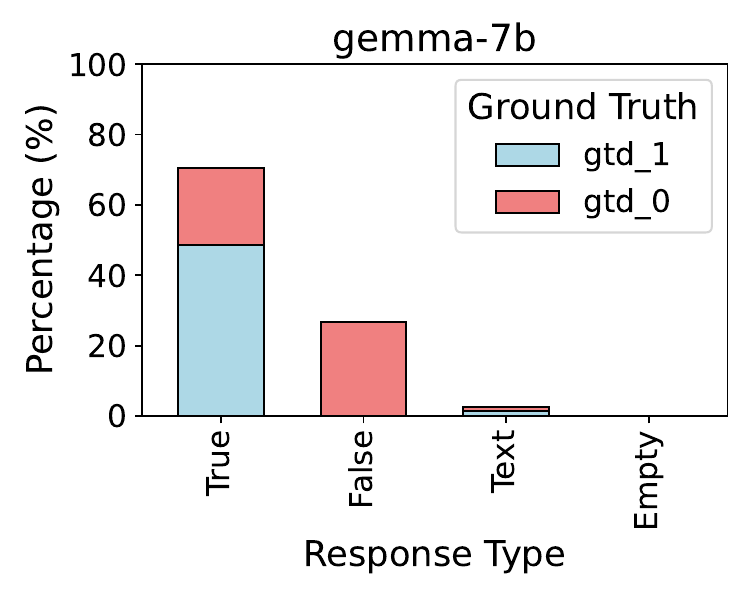} \\
    
    \includegraphics[width=0.22\linewidth]{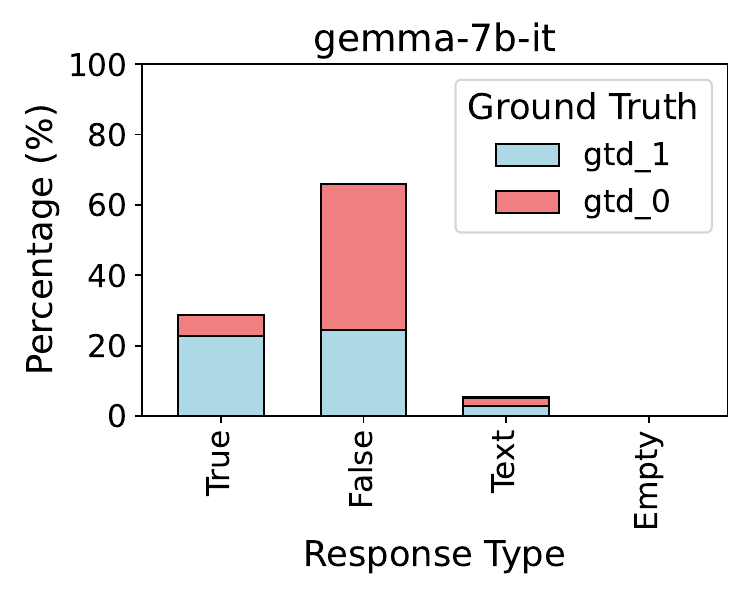} \hfill
    \includegraphics[width=0.22\linewidth]{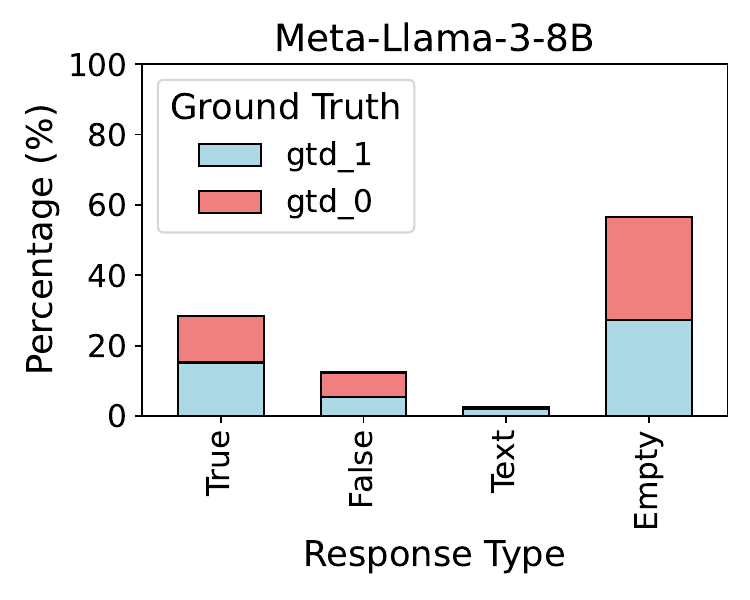} \hfill
    \includegraphics[width=0.22\linewidth]{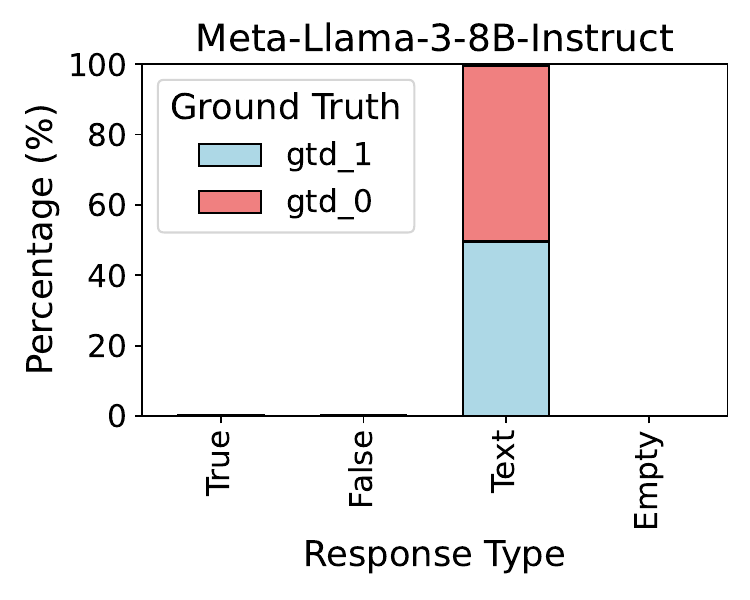} \hfill
    \includegraphics[width=0.22\linewidth]{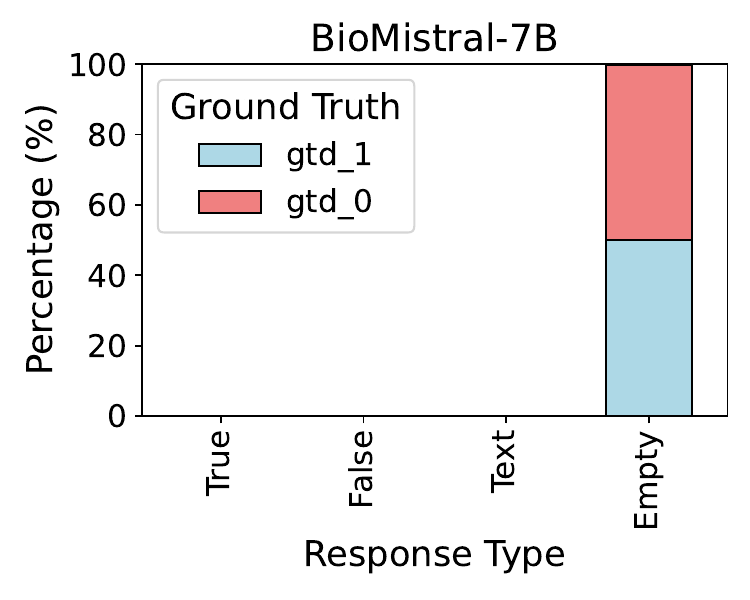}
    
    \caption{Percentage distribution of model response types under few-shot settings for prompts with all distractors for the set of biologically factual argumentative texts.}
    \label{fig:eight_plots_all_ICL}
\end{figure*}

We observed four types of text outputs: those aligned with the instruction (regardless of correctness), empty outputs where no text was generated, incorrect text outputs such as repeated prompts or random content, and outputs resembling Chain-of-Thought (CoT) reasoning that, while potentially containing correct reasoning, did not align with the given instructions (Figs. \ref{fig:eight_plots_zeroshot}-\ref{fig:eight_plots_all_ICL}). 
We noticed that BioMistral-7B generated empty outputs in 100\% of the cases regardless of the specific settings, while Meta-Llama-3-8B exhibits this behaviour for ZS settings in both tasks. We attribute this observation to safety mechanisms applied during pre-training (\cite{labrak2024biomistral}), suggesting that domain-specific knowledge about human genome pathways is absent in both models. Similarly, Mistral-7B-v0.1 responses simply repeat the prompt text in 88\% of the cases in the ZS settings, and 69\% of the cases in FS (Table \ref{tab:response_type_ndistr_0}).
Moreover, CoT outputs including phrases like e.g. "A nice logical puzzle! Let's break it down step by step..." were particularly common for Meta-Llama-3-8B Instruct, which often ignored the specific instructions to address the task. This behaviour highlights potential biases introduced during instruction-tuning which make the models unable to generalise to domains that are out-of-distribution of the training set.

\section{Results: Ambiguous Impact of Distractors on Reasoning}
\label{sec:appendix_results_distractors}
\begin{figure*}[htbp]
\centering
\includegraphics[width= .99\textwidth]{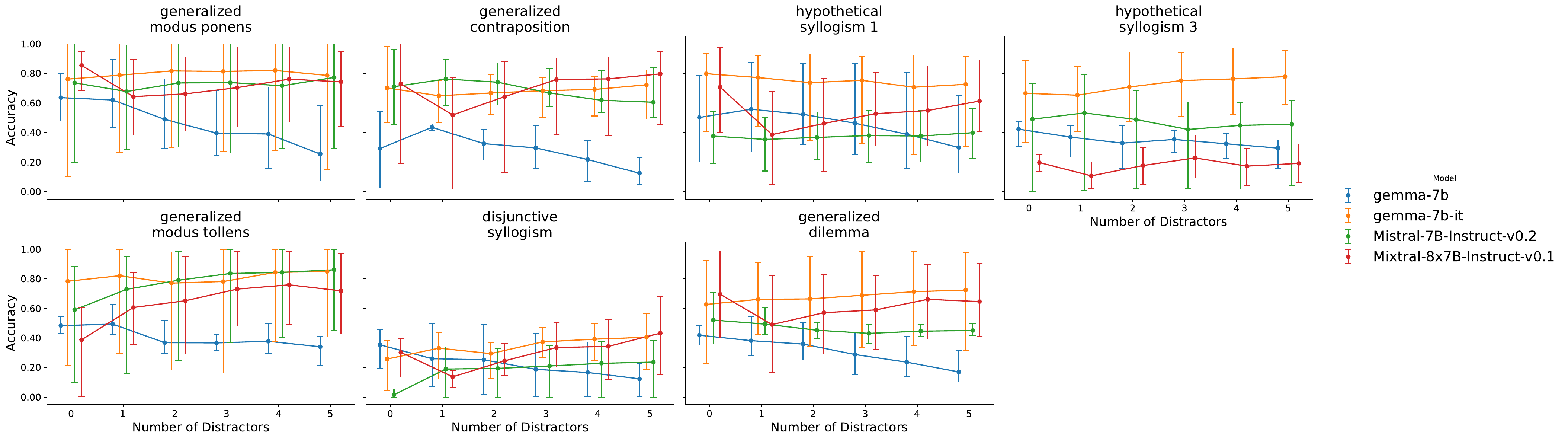}
\caption{Task 1: Accuracy versus number of distractors and scheme in the zero-shot setting. Lines connect the average values for each model, with error bars representing the range (min-max).}
\label{fig:task1_bydistractors_zeroshot}
\end{figure*}

\begin{figure*}[htbp]
\centering
\includegraphics[width= .99\textwidth]{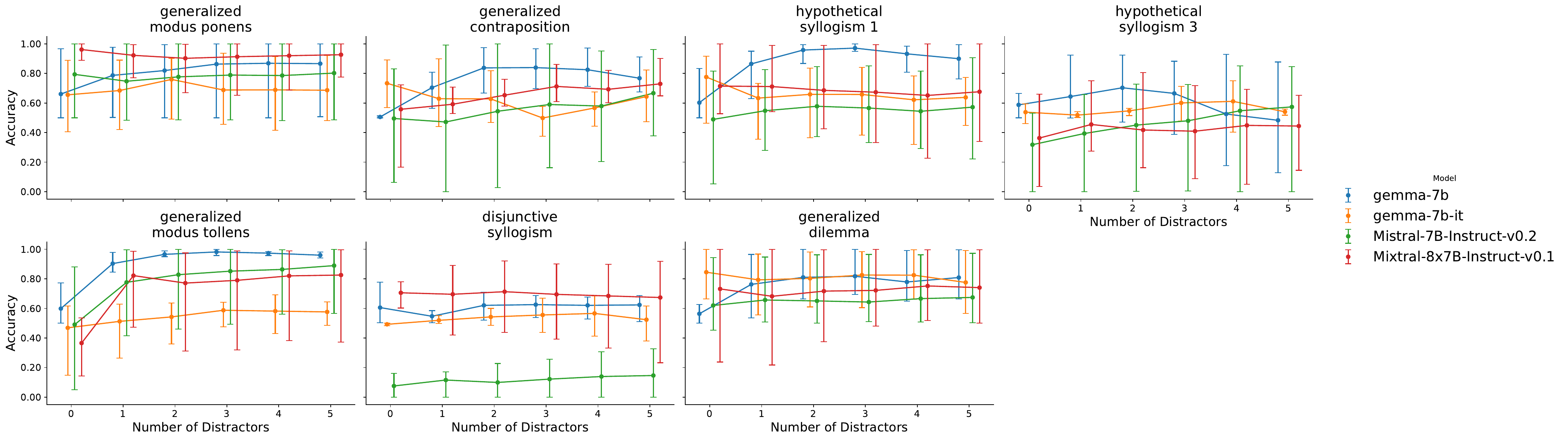}
\caption{Task 1: Accuracy versus number of distractors and scheme in the few-shot setting. Lines connect the average values for each model, with error bars representing the range (min-max).}
\label{fig:task1_bydistractors_icl}
\end{figure*}

\begin{figure*}[htbp]
\centering
\includegraphics[width= .99\textwidth]{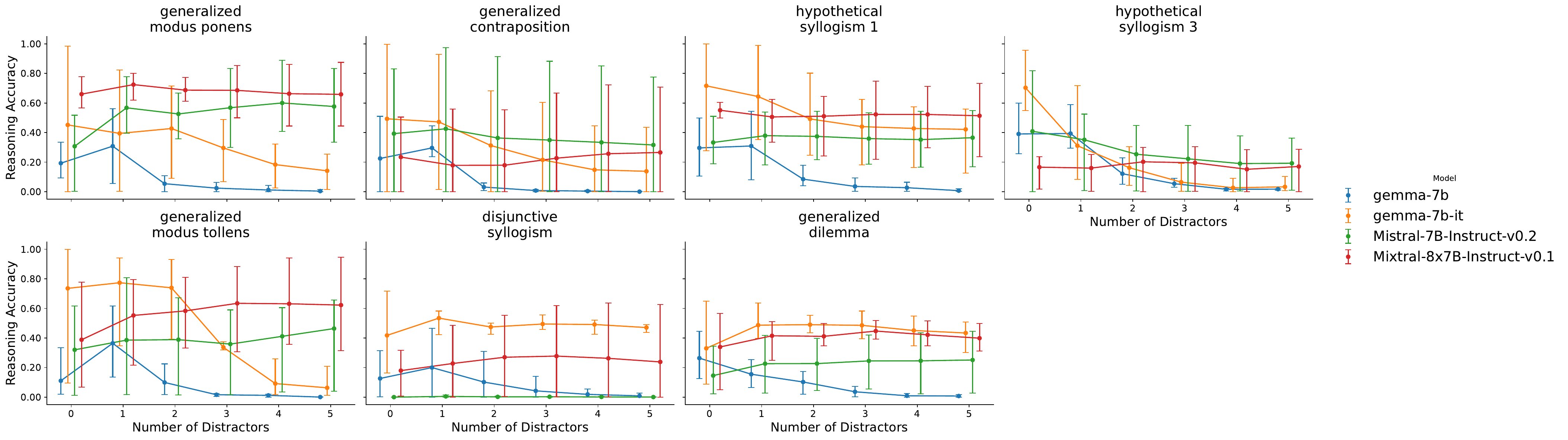}
\caption{Task 2: Reasoning accuracy versus number of distractors and scheme in the zero-shot setting. Lines connect the average values for each model, with error bars representing the range (min-max).}
\label{fig:task2_bydistractors_zeroshot}
\end{figure*}

\begin{figure*}[htbp]
\centering
\includegraphics[width= .99\textwidth]{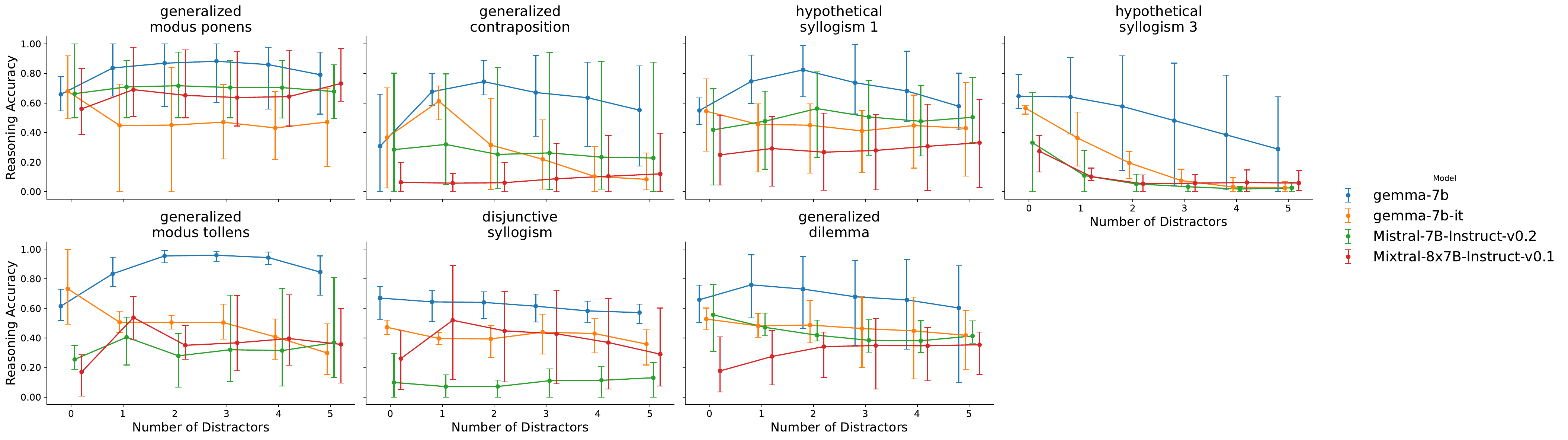}
\caption{Task 2: Reasoning accuracy versus number of distractors and scheme in the few-shot setting. Lines connect the average values for each model, with error bars representing the range (min-max).}
\label{fig:task2_bydistractors_icl}
\end{figure*}

LLMs show a slight sensitivity to increasing number of distractors $n_D$ in the prompt, with overall accuracy remaining stable (Figs. \ref{fig:task1_bydistractors_zeroshot}- \ref{fig:task2_bydistractors_icl}, Table \ref{tab:response_type_ndistr_all}).

While some models struggle with increasing $n_D$, others can leverage few-shot learning to mitigate their impact, though the effect is scheme-dependent.
Considering Task 1, in the ZS setting (Fig. \ref{fig:task1_bydistractors_zeroshot}, Table \ref{tab:correlation_by_distractors_task1}), Gemma-7b shows a significant decline in accuracy as $n_D$ increases, particularly in the \textit{generalized dilemma} ($r = -0.643, p = 0.001$), \textit{generalized modus ponens} ($r=-0.592, p = 0.002$), and \textit{generalized modus tollens} ($r = -0.571, p = 0.004$) schemes, indicating a moderate negative correlation. In contrast, in the ZS setting, Mistral-7B Instruct-v0.2 exibits a moderate improvement in accuracy with higher $n_D$ , in the \textit{generalized modus tollens} ($r = 0.540, p = 0.006$) scheme, reflecting a weak positive correlation overall ($r = 0.333, p < 0.001$). 
Considering \textit{reasoning accuracy}, in the ZS setting (Fig. \ref{fig:task2_bydistractors_zeroshot}, Table \ref{tab:correlation_by_distractors_task2}), the Gemma-7b model exhibited a substantial drop as the $n_D$ increased ($r=-0.951, p<0.001$), with an initial low accuracy of 0.3 even with $n_D=0$ . The steepest declines were observed in \textit{hypothetical syllogism 1} ($r=-1.0, p<0.000$) and \textit{generalized dilemma} ($r=-1.0$, $p<0.000$) schemes. For the Gemma-7b-it, the strongest negative correlation between the model and $n_D$ was for the \textit{hypothetical syllogism 1}, \textit{generalized modus ponens} and \textit{generalized contraposition} schemes ($r=-1.0, p=0.000$).
In the FS setting (Fig. \ref{fig:task2_bydistractors_icl}), the Gemma-7b-it model consistently exhibited significant decreases in \textit{reasoning accuracy} across all schemes, with the most pronounced effect in \textit{hypothetical syllogism 3} and \textit{generalized modus tollens} ($r=-1.0 p<0.000$). 
Interestingly, the Mistral-7B Instruct model in both settings, depending on the scheme, showed a positive or negative significant correlation.

The findings underscore the substantial impact of distractors on reasoning accuracy, particularly in complex syllogistic reasoning tasks, revealing that current LLMs are highly susceptible to performance degradation as distractor complexity increases.

\section{Results: Models Prioritize Contextual Knowledge Over Background Knowledge}
\label{sec:appendix_results_dummy}
\begin{figure*}[t]
  \centering
  \begin{tabular}{cc}
    \includegraphics[width=0.4\linewidth]{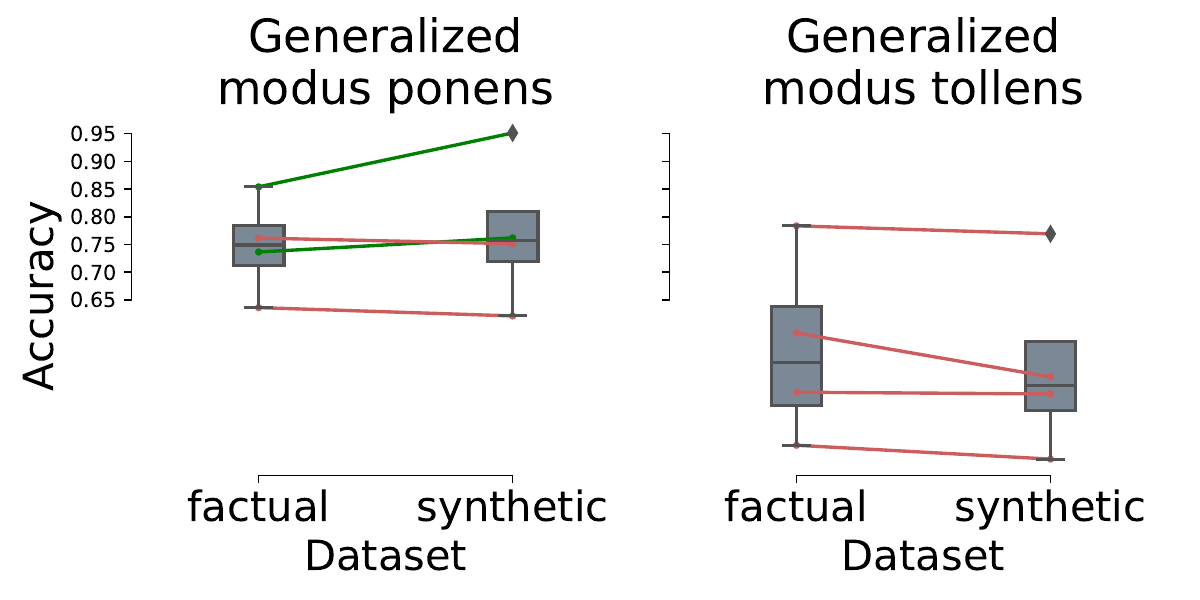} & 
    \includegraphics[width=0.4\linewidth]{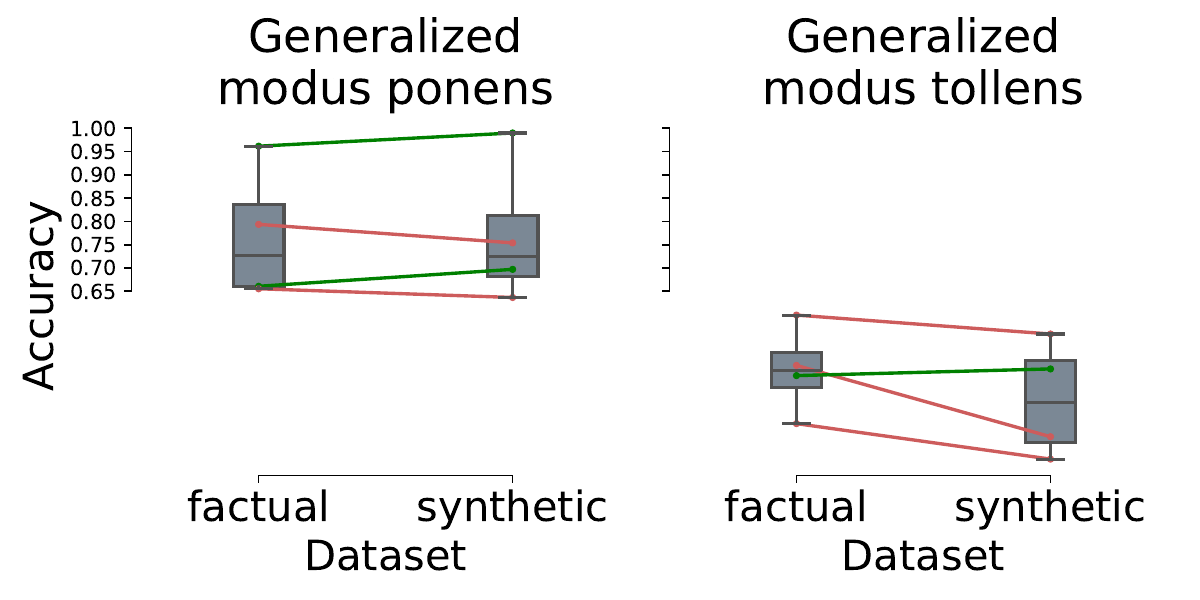} \\
    \includegraphics[width=0.4\linewidth]{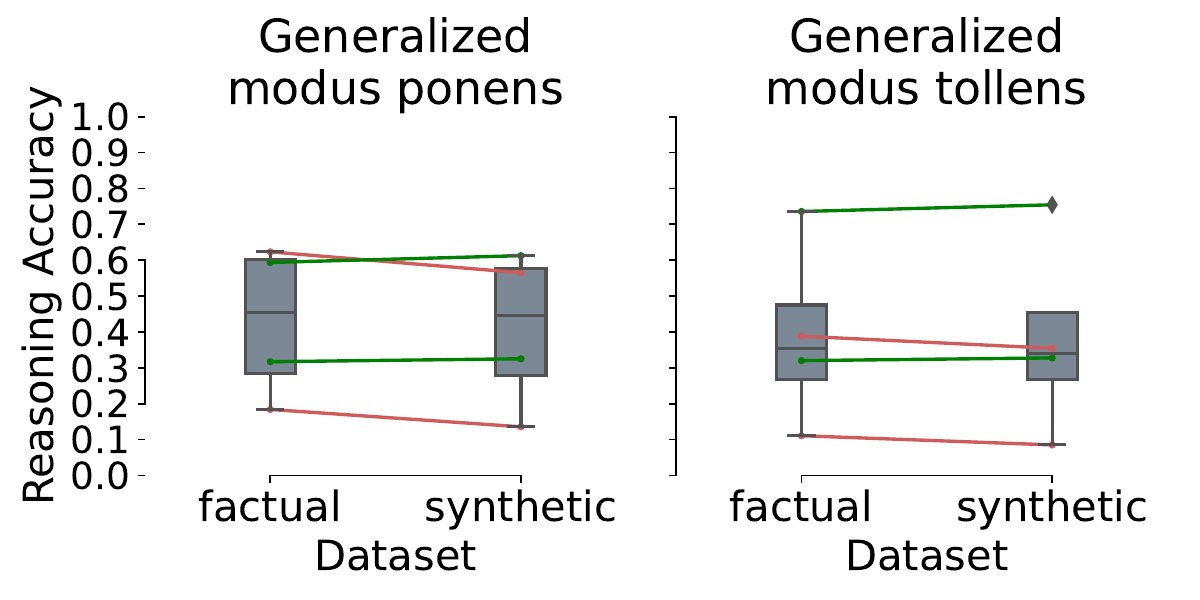} & 
    \includegraphics[width=0.4\linewidth]{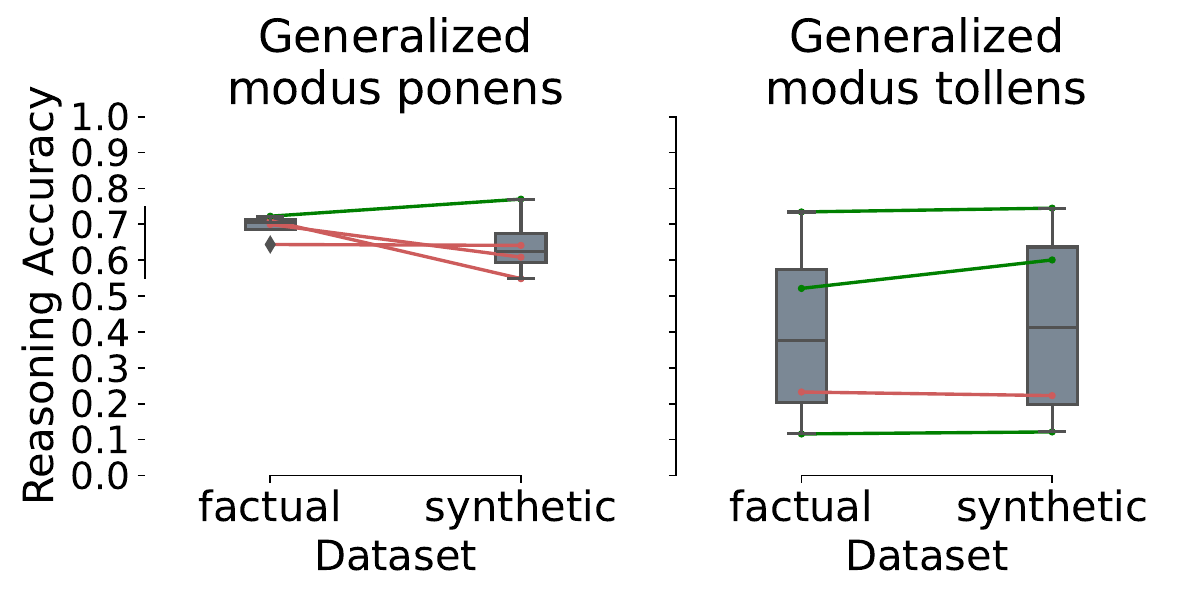} \\
  \end{tabular}
   \caption {Accuracy comparison between two datasets: Biologically Factual vs. Artificial with Synthetic Names for Task 1 (top) and Task 2 (bottom); ZS (left) and FS (right). Lines connect the accuracy for each model, with green indicating an increase and red indicating a decrease. Gray boxplots display the median, Q1, Q3, and the range (minimum to maximum) of the data.}
  \label{fig:combined_paired_plot_comparison}
\end{figure*}

The lack of statistically significant differences (Fig. \ref{fig:combined_paired_plot_comparison}) in accuracy between biologically factual and artificial datasets across \textit{generalized modus ponens} and \textit{generalized modus tollens} schemes suggests that the models' reasoning capabilities rely more on stated contextual knowledge and logical structure than on pre-existing background knowledge. This holds true for both \textit{accuracy} and \textit{reasoning accuracy}, as well as in both ZS and FS settings: models that perform well on a given scheme maintain their performance even when factual gene names are replaced by synthetic names, and the same consistency is observed for models with weaker performance. This ability to maintain accuracy with synthetic gene names in the artificial set demonstrates that models can abstract and apply logical reasoning independently of their internal domain-specific knowledge.

\section{Supplementary Figures}
\label{sec:appendix_figures}

\section{Supplementary Tables }
\label{sec:appendix_tables}


\begin{table*}[t]
  \centering
  \resizebox{1.0\textwidth}{!}{%
  \begin{tabular}{lllllllll|llllllll}
    \toprule
    \textbf{Model} & \multicolumn{8}{c}{\textbf{ZS}} & \multicolumn{8}{c}{\textbf{FS}} \\
     & \begin{tabular}[c]{@{}l@{}}Non-empty \\ output\end{tabular} & \begin{tabular}[c]{@{}l@{}}Irrelevant \\ text\end{tabular} & \begin{tabular}[c]{@{}l@{}}Following \\ instruction\end{tabular} & \textbf{Acc.} & \textbf{Recall} & \textbf{Precision} & \textbf{F1} & \textbf{Faith.} & \begin{tabular}[c]{@{}l@{}}Non-empty \\ output\end{tabular} & \begin{tabular}[c]{@{}l@{}}Irrelevant \\ text\end{tabular} & \begin{tabular}[c]{@{}l@{}}Following \\ instruction\end{tabular} & \textbf{Acc.} & \textbf{Recall} & \textbf{Precision} & \textbf{F1} & \textbf{Faith.} \\
    \toprule
    BioMistral-7B & 0.00 & - & - & - & - & - & - & - & 0.00 & - & - & - & - & - & - & - \\ \midrule
    Meta-Llama-3-8B & 0.01 & 0.01 & 0.00 & 0.00 & 0.00 & 0.00 & 0.00      & 0.00 & 0.44 & 0.06 & 0.38 & 0.19 & 0.25 & 0.23 & 0.24 & 0.08 \\
    Meta-Llama-3-8B-Instruct   & 1.00              & 0.93         & 0.07            & 0.06     & 0.07   & 0.06      & 0.06      &         0.01     & 1.00              & 0.99         & 0.01            & 0.00     & 0.00   & 0.00      & 0.00      &           0.00   \\ \midrule
    Mistral-7B-Instruct-v0.2   & 1.00              & 0.08          & 0.92           & 0.51     & 0.70   & 0.51      & 0.59      &       0.43       & 1.00              & 0.18         & 0.82           & 0.54     & 0.60   & 0.53      & 0.57      &             0.53 \\
    Mistral-7B-v0.1            & 0.33               & 0.33         & 0.00            & 0.00     & 0.00   & 0.00      & 0.00      &    0.00          & 0.17               & 0.16         & 0.01            & 0.00     & 0.00   & 0.00      & 0.00      &         0.00     \\
    Mixtral-8x7B-Instruct-v0.1 & 1.00              & 0.37         & 0.63           & 0.51     & 0.40   & 0.51      & 0.45      &    0.33          & 1.00              & 0.06          & 0.94           & 0.68     & 0.62   & 0.70      & 0.66      &          0.56    \\ \midrule
    Gemma-7b                   & 0.79               & 0.15         & 0.64           & 0.34     & 0.56   & 0.39      & 0.46      &       0.09       & 0.99               & 0.02          & 0.97           & 0.75     & 0.97   & 0.68      & 0.80      &          0.53    \\
    Gemma-7b-it                & 1.00              & 0.00          & 1.00          & 0.67     & 0.73   & 0.65      & 0.69      &      0.63        & 1.00              & 0.08          & 0.92           & 0.62     & 0.43   & 0.70      & 0.53      &       0.39       \\
    \bottomrule\\
  \end{tabular}
  }
  \caption{Task 1: Distribution of model response types and performance outcomes across two experimental settings—ZS and FS, considering all distractor conditions (n distractors from 0 to 5), for all syllogistic schemes within biologically factual argumentative texts. Response types include: non-empty outputs, irrelevant text generation, and outputs adhering to the given instructions.
  }
  \label{tab:response_type_ndistr_all}
\end{table*}

\begin{table}[]
\centering
\resizebox{0.5\columnwidth}{!}{%
\begin{tabular}{llllllllll}
\hline
\multirow{2}{*}{\textbf{Scheme}}                                                               & \multirow{2}{*}{\textbf{Model}} & \multicolumn{2}{l}{\textbf{Recall}} & \multicolumn{2}{l}{\textbf{Precision}}    & \multicolumn{2}{l}{\textbf{F1-score}}  & \multicolumn{2}{l}{\textbf{Accuracy}}  \\ \cline{3-10} 
                                                                                               &                                 & \textbf{ZS} & \textbf{FS}      & \textbf{ZS} & \textbf{FS}      & \textbf{ZS} & \textbf{FS}      & \textbf{ZS} & \textbf{FS}      \\ \hline
\multirow{9}{*}{\textbf{\begin{tabular}[c]{@{}l@{}}disjunctive \\ syllogism\end{tabular}}}      & All models                      & 0.11                & 0.24          & 0.11                  & 0.25           & 0.11                 & 0.25           & 0.12                 & 0.27           \\
                                                                                                & BioMistral-7B                   & 0.00                & 0.00          & 0.00                  & 0.00           & 0.00                 & 0.00           & 0.00                 & 0.00           \\
                                                                                                & Meta-Llama-3-8B                 & 0.01                & 0.43          & 0.01                  & 0.33           & 0.01                 & 0.38           & 0.01                 & 0.29           \\
                                                                                                & Meta-Llama-3-8B-Instruct        & 0.00                & 0.00          & 0.00                  & 0.00           & 0.00                 & 0.00           & 0.00                 & 0.00           \\
                                                                                                & Mistral-7B-Instruct-v0.2        & 0.03                & 0.03          & 0.03                  & 0.03           & 0.03                 & 0.03           & 0.02        & 0.08  \\
                                                                                                & Mistral-7B-v0.1                 & 0.00                & 0.00          & 0.00                  & 0.00           & 0.00                 & 0.00           & 0.00                 & 0.00           \\
                                                                                                & Mixtral-8x7B-Instruct-v0.1      & 0.11                & 0.42          & 0.18                  & 0.98           & 0.13                 & 0.59           & 0.30                 & \textbf{0.71}           \\
                                                                                                & Gemma-7b                        & 0.65                & 1.00          & 0.41                  & 0.56           & 0.50                 & 0.72           & \textbf{0.35}                 & 0.61           \\
                                                                                                & Gemma-7b-it                     & 0.08                & 0.02          & 0.13                  & 0.34           & 0.10                 & 0.03           & 0.26        & 0.49  \\ \midrule
\multirow{9}{*}{\textbf{\begin{tabular}[c]{@{}l@{}}generalized \\ contraposition\end{tabular}}} & All models                      & 0.39                & 0.38          & 0.33                  & 0.34           & 0.35                 & 0.36           & 0.30                 & 0.31           \\
                                                                                                & BioMistral-7B                   & 0.00                & 0.00          & 0.00                  & 0.00           & 0.00                 & 0.00           & 0.00                 & 0.00           \\
                                                                                                & Meta-Llama-3-8B                 & 0.01                & 0.46          & 0.01                  & 0.35           & 0.01                 & 0.39           & 0.01                 & 0.29           \\
                                                                                                & Meta-Llama-3-8B-Instruct        & 0.10                & 0.00          & 0.09                  & 0.00           & 0.10                 & 0.00           & 0.06                 & 0.00           \\
                                                                                                & Mistral-7B-Instruct-v0.2        & 0.97                & 0.39          & 0.61                  & 0.44           & 0.75                 & 0.41           & 0.67                 & 0.45           \\
                                                                                                & Mistral-7B-v0.1                 & 0.00                & 0.1          & 0.00                  & 0.01           & 0.00                 & 0.01           & 0.00                 & 0.00           \\
                                                                                                & Mixtral-8x7B-Instruct-v0.1      & 0.61                & 0.45          & 0.72                  & 0.55           & 0.67                 & 0.49           & \textbf{0.69}                 & 0.54           \\
                                                                                                & Gemma-7b                        & 0.42                & 1.00          & 0.32                  & 0.50           & 0.36                 & 0.67           & 0.26        & 0.51  \\
                                                                                                & Gemma-7b-it                     & 0.98                & 0.74          & 0.61                  & 0.71           & 0.75                 & 0.73           & 0.68                 & \textbf{0.72}           \\ \midrule
\multirow{9}{*}{\textbf{\begin{tabular}[c]{@{}l@{}}generalized \\ dilemma\end{tabular}}}        & All models                      & 0.37                & 0.48          & 0.32                  & 0.41           & 0.34                 & 0.45           & 0.28                 & 0.40           \\
                                                                                                & BioMistral-7B                   & 0.00                & 0.00          & 0.00                  & 0.00           & 0.00                 & 0.00           & 0.00                 & 0.00           \\
                                                                                                & Meta-Llama-3-8B                 & 0.01                & 0.61          & 0.01                  & 0.45           & 0.01                 & 0.52           & 0.00                 & 0.43           \\
                                                                                                & Meta-Llama-3-8B-Instruct        & 0.00                & 0.00          & 0.00                  & 0.00           & 0.00                 & 0.00           & 0.00                 & 0.00           \\
                                                                                                & Mistral-7B-Instruct-v0.2        & 0.91                & 0.86          & 0.51                  & 0.58           & 0.65                 & 0.69           & 0.52                 & 0.62           \\
                                                                                                & Mistral-7B-v0.1                 & 0.00                & 0.00          & 0.00                  & 0.00           & 0.00                 & 0.00           & 0.00                 & 0.00           \\
                                                                                                & Mixtral-8x7B-Instruct-v0.1      & 0.48                & 0.63          & 0.85                  & 0.79           & 0.61                 & 0.70           & \textbf{0.70}                 & 0.73           \\
                                                                                                & Gemma-7b                        & 0.79                & 1.00          & 0.45                  & 0.53           & 0.58                 & 0.70           & 0.42                 & 0.56           \\
                                                                                                & Gemma-7b-it                     & 0.82                & 0.79          & 0.59                  & 0.89           & 0.69                 & 0.84           & 0.63                 & \textbf{0.85}  \\ \midrule
\multirow{9}{*}{\textbf{\begin{tabular}[c]{@{}l@{}}generalized \\ modus ponens\end{tabular}}}   & All models                      & 0.51                & 0.50          & 0.43                  & 0.45           & 0.47                 & 0.47           & 0.41                 & 0.44           \\
                                                                                                & BioMistral-7B                   & 0.00                & 0.00          & 0.00                  & 0.00           & 0.00                 & 0.00           & 0.00                 & 0.00           \\
                                                                                                & Meta-Llama-3-8B                 & 0.00                & 0.59          & 0.00                  & 0.44           & 0.00                 & 0.50           & 0.00                 & 0.41           \\
                                                                                                & Meta-Llama-3-8B-Instruct        & 0.75                & 0.03          & 0.48                  & 0.03           & 0.59                 & 0.03           & 0.47                 & 0.02           \\
                                                                                                & Mistral-7B-Instruct-v0.2        & 0.80                & 0.99          & 0.65                  & 0.73           & 0.72                 & 0.84           & 0.69        & 0.82  \\
                                                                                                & Mistral-7B-v0.1                 & 0.00                & 0.00          & 0.00                  & 0.00           & 0.00                 & 0.00           & 0.00                 & 0.00           \\
                                                                                                & Mixtral-8x7B-Instruct-v0.1      & 0.91                & 1.00          & 0.80                  & 0.97           & 0.85                 & 0.98           & \textbf{0.84}        & \textbf{0.98}  \\
                                                                                                & Gemma-7b                        & 0.91                & 1.00          & 0.56                  & 0.63           & 0.69                 & 0.78           & 0.59        & 0.71  \\
                                                                                                & Gemma-7b-it                     & 0.74                & 0.37          & 0.69                  & 0.64           & 0.71                 & 0.47           & 0.70                 & 0.58           \\ \midrule
\multirow{9}{*}{\textbf{\begin{tabular}[c]{@{}l@{}}generalized \\ modus tollens\end{tabular}}}  & All models                      & 0.40                & 0.28          & 0.34                  & 0.28           & 0.36                 & 0.28           & 0.31                 & 0.29           \\
                                                                                                & BioMistral-7B                   & 0.00                & 0.00          & 0.00                  & 0.00           & 0.00                 & 0.00           & 0.00                 & 0.00           \\
                                                                                                & Meta-Llama-3-8B                 & 0.01                & 0.48          & 0.01                  & 0.40           & 0.01                 & 0.43           & 0.01                 & 0.37           \\
                                                                                                & Meta-Llama-3-8B-Instruct        & 0.32                & 0.00          & 0.26                  & 0.00           & 0.28                 & 0.00           & 0.20                 & 0.00           \\
                                                                                                & Mistral-7B-Instruct-v0.2        & 0.70                & 0.42          & 0.57                  & 0.49           & 0.63                 & 0.45           & 0.59                 & 0.49           \\
                                                                                                & Mistral-7B-v0.1                 & 0.00                & 0.00          & 0.00                  & 0.00           & 0.00                 & 0.00           & 0.00                 & 0.00           \\
                                                                                                & Mixtral-8x7B-Instruct-v0.1      & 0.50                & 0.14          & 0.41                  & 0.26           & 0.45                 & 0.18           & 0.39                 & 0.37  \\
                                                                                                & Gemma-7b                        & 0.92                & 1.00          & 0.49                  & 0.55           & 0.64                 & 0.71           & 0.48                 & \textbf{0.60}           \\
                                                                                                & Gemma-7b-it                     & 0.71                & 0.17          & 0.83                  & 0.42           & 0.77                 & 0.24           & \textbf{0.78}        & 0.47           \\ \midrule
\multirow{9}{*}{\textbf{\begin{tabular}[c]{@{}l@{}}hypothetical \\ syllogism 1\end{tabular}}}   & All models                      & 0.36                & 0.37          & 0.32                  & 0.36           & 0.34                 & 0.37           & 0.30                 & 0.36           \\
                                                                                                & BioMistral-7B                   & 0.00                & 0.00          & 0.00                  & 0.00           & 0.00                 & 0.00           & 0.00                 & 0.00           \\
                                                                                                & Meta-Llama-3-8B                 & 0.01                & 0.54          & 0.01                  & 0.42           & 0.01                 & 0.47           & 0.01                 & 0.40           \\
                                                                                                & Meta-Llama-3-8B-Instruct        & 0.16                & 0.00          & 0.14                  & 0.00           & 0.15                 & 0.00           & 0.10                 & 0.00           \\
                                                                                                & Mistral-7B-Instruct-v0.2        & 0.59                & 0.42          & 0.40                  & 0.44           & 0.48                 & 0.43           & 0.35                 & 0.44           \\
                                                                                                & Mistral-7B-v0.1                 & 0.00                & 0.01          & 0.00                  & 0.01           & 0.00                 & 0.01           & 0.00                 & 0.01           \\
                                                                                                & Mixtral-8x7B-Instruct-v0.1      & 0.55                & 0.36          & 0.72                  & 0.97           & 0.63                 & 0.53           & 0.67                 & 0.67           \\
                                                                                                & Gemma-7b                        & 0.74                & 1.00          & 0.48                  & 0.56           & 0.58                 & 0.72           & 0.46                 & 0.61           \\
                                                                                                & Gemma-7b-it                     & 0.80                & 0.62          & 0.77                  & 0.85           & 0.78                 & 0.72           & \textbf{0.78}        & \textbf{0.76}           \\ \midrule
\multirow{9}{*}{\textbf{\begin{tabular}[c]{@{}l@{}}hypothetical \\ syllogism 3\end{tabular}}}   & All models                      & 0.30                & 0.29          & 0.26                  & 0.28           & 0.28                 & 0.28           & 0.23                 & 0.27           \\
                                                                                                & BioMistral-7B                   & 0.00                & 0.00          & 0.00                  & 0.00           & 0.00                 & 0.00           & 0.00                 & 0.00           \\
                                                                                                & Meta-Llama-3-8B                 & 0.00                & 0.45          & 0.00                  & 0.34           & 0.00                 & 0.39           & 0.00                 & 0.29           \\
                                                                                                & Meta-Llama-3-8B-Instruct        & 0.04                & 0.00          & 0.03                  & 0.00           & 0.03                 & 0.00           & 0.03                 & 0.00           \\
                                                                                                & Mistral-7B-Instruct-v0.2        & 0.48                & 0.13          & 0.49                  & 0.21           & 0.49                 & 0.16           & 0.49                 & 0.32           \\
                                                                                                & Mistral-7B-v0.1                 & 0.00                & 0.00          & 0.00                  & 0.00           & 0.00                 & 0.00           & 0.00                 & 0.00           \\
                                                                                                & Mixtral-8x7B-Instruct-v0.1      & 0.15                & 0.23          & 0.16                  & 0.31           & 0.15                 & 0.27           & 0.20        & 0.36           \\
                                                                                                & Gemma-7b                        & 0.84                & 1.00          & 0.46                  & 0.55           & 0.59                 & 0.71           & 0.42                 & \textbf{0.59}           \\
                                                                                                & Gemma-7b-it                     & 0.87                & 0.54          & 0.62                  & 0.54           & 0.72                 & 0.54           & \textbf{0.67}                 & 0.54           \\ \bottomrule
\end{tabular}%
}
\caption{Results from Task 1 baseline models on the biologically factual argumentative texts set (without synthetic data and without distractors, bold - the best and worst accuracy values for each model).}
\label{tab:task1_byscheme}
\end{table}

\begin{table}[]
\centering
\resizebox{0.5\columnwidth}{!}{%
\begin{tabular}{@{}lllllllll@{}}
\toprule
\multirow{2}{*}{Scheme}                                              & \multicolumn{2}{l}{Gemma-7b}     & \multicolumn{2}{l}{Gemma-7b-it} & \multicolumn{2}{l}{\begin{tabular}[c]{@{}l@{}}Mistral-7B\\ Instruct-v0.2\end{tabular}} & \multicolumn{2}{l}{\begin{tabular}[c]{@{}l@{}}Mixtral-8x7B\\ Instruct-v0.1\end{tabular}} \\ \cmidrule(l){2-9} 
                                                                     & r               & p-val          & r              & p-val          & r                                          & p-val                                     & r                                           & p-val                                      \\ \midrule
\multicolumn{9}{l}{ZS}                                                                                                                                                                                                                                                                                                 \\ \midrule
\begin{tabular}[c]{@{}l@{}}disjunctive\\ syllogism\end{tabular}      & \textbf{-0.497} & \textbf{0.013} & \textbf{0.432} & \textbf{0.035} & \textbf{0.405}                                      & \textbf{0.050}                                     & \textbf{0.418}                              & \textbf{0.042}                             \\
\begin{tabular}[c]{@{}l@{}}generalized\\ contraposition\end{tabular} & \textbf{-0.495} & \textbf{0.014} & 0.155          & 0.469          & -0.397                                     & 0.055                                     & 0.229                                       & 0.282                                      \\
\begin{tabular}[c]{@{}l@{}}generalized\\ dilemma\end{tabular}        & \textbf{-0.643} & \textbf{0.001} & 0.203          & 0.342          & -0.282                                     & 0.182                                     & 0.049                                       & 0.819                                      \\
\begin{tabular}[c]{@{}l@{}}generalized\\ modus ponens\end{tabular}   & \textbf{-0.592} & \textbf{0.002} & 0.104          & 0.630          & 0.069                                      & 0.748                                     & -0.026                                      & 0.902                                      \\
\begin{tabular}[c]{@{}l@{}}generalized\\ modus tollens\end{tabular}  & \textbf{-0.571} & \textbf{0.004} & 0.198          & 0.353          & \textbf{0.540}                             & \textbf{0.006}                            & \textbf{0.432}                              & \textbf{0.035}                             \\
\begin{tabular}[c]{@{}l@{}}hypothetical\\ syllogism 1\end{tabular}   & -0.375          & 0.071          & -0.268         & 0.205          & 0.115                                      & 0.594                                     & 0.012                                       & 0.954                                      \\
\begin{tabular}[c]{@{}l@{}}hypothetical\\ syllogism 3\end{tabular}   & \textbf{-0.465} & \textbf{0.022} & 0.211          & 0.321          & -0.125                                     & 0.560                                     & 0.100                                       & 0.640                                      \\ \midrule
\multicolumn{9}{l}{FS}                                                                                                                                                                                                                                                                                                       \\ \midrule
\begin{tabular}[c]{@{}l@{}}disjunctive\\ syllogism\end{tabular}      & 0.279 & 0.187 & 0.370 & 0.075 & 0.197                             & 0.356                            & 0.116                                       & 0.588                                      \\
\begin{tabular}[c]{@{}l@{}}generalized\\ contraposition\end{tabular} & \textbf{0.511}           & \textbf{0.011}          & -0.201         & 0.347          & 0.189                                      & 0.377                                     & 0.349                                       & 0.095                                      \\
\begin{tabular}[c]{@{}l@{}}generalized\\ dilemma\end{tabular}        & \textbf{0.462}          & \textbf{0.023}          & -0.097         & 0.652          & 0.183                                      & 0.391                                     & -0.046                                      & 0.831                                      \\
\begin{tabular}[c]{@{}l@{}}generalized\\ modus ponens\end{tabular}   & \textbf{0.455}           & \textbf{0.025}          & 0.051         & 0.812          & 0.011                                     & 0.961                                     & 0.023                                      & 0.914                                      \\
\begin{tabular}[c]{@{}l@{}}generalized\\ modus tollens\end{tabular}  & \textbf{0.568}  & \textbf{0.004} & 0.224          & 0.292          & \textbf{0.419}                                      & \textbf{0.041}                                     & 0.391                              & 0.059                             \\
\begin{tabular}[c]{@{}l@{}}hypothetical\\ syllogism 1\end{tabular}   & \textbf{0.420}           & \textbf{0.041}          & -0.314         & 0.135          & 0.056                                      & 0.793                                     & -0.026                                      & 0.902                                      \\
\begin{tabular}[c]{@{}l@{}}hypothetical\\ syllogism 3\end{tabular}   & -0.122 & 0.571 & 0.199          & 0.351          & \textbf{0.422}                             & \textbf{0.040}                            & 0.116                              & 0.588                             \\ \bottomrule
\end{tabular}%
}
\caption{Values of the Spearman’s Ranked Correlation Coefficient (r) for Accuracy by Distractors and Syllogistic Scheme for evaluated Models in Task 1: Spearman's correlation coefficients (r) and p-values for the accuracy metric are shown across various levels of distractors and syllogistic schemes for each model. Negative r values reflect a decrease in accuracy with increasing distractor complexity. The highest correlation values for each scheme are highlighted in bold, indicating the models most affected by distractors.}
\label{tab:correlation_by_distractors_task1}
\end{table}

\begin{table}[]
\resizebox{\columnwidth}{!}{%
\begin{tabular}{@{}lllllllllllllllllll@{}}
\toprule
\multirow{2}{*}{\textbf{Scheme}}                                      & \multicolumn{2}{l}{\textbf{BioMistral-7B}} & \multicolumn{2}{l}{\textbf{Meta-Llama-3-8B}} & \multicolumn{2}{l}{\textbf{\begin{tabular}[c]{@{}l@{}}Meta-Llama-3-8B\\ Instruct\end{tabular}}} & \multicolumn{2}{l}{\textbf{\begin{tabular}[c]{@{}l@{}}Mistral-7B\\ Instruct-v0.2\end{tabular}}} & \multicolumn{2}{l}{\textbf{Mistral-7B-v0.1}} & \multicolumn{2}{l}{\textbf{\begin{tabular}[c]{@{}l@{}}Mixtral-8x7B\\ Instruct-v0.1\end{tabular}}} & \multicolumn{2}{l}{\textbf{Gemma-7b}} & \multicolumn{2}{l}{\textbf{Gemma-7b-it}} & \multicolumn{2}{l}{\textbf{All models}} \\ \cmidrule(l){2-19} 
                                                                      & \textbf{ZS}      & \textbf{FS}     & \textbf{ZS}       & \textbf{FS}      & \textbf{ZS}                                & \textbf{FS}                                & \textbf{ZS}                                & \textbf{FS}                                & \textbf{ZS}       & \textbf{FS}      & \textbf{ZS}                                 & \textbf{FS}                                 & \textbf{ZS}   & \textbf{FS}   & \textbf{ZS}    & \textbf{FS}     & \textbf{ZS}    & \textbf{FS}    \\ \midrule
\begin{tabular}[c]{@{}l@{}}disjunctive \\ syllogism\end{tabular}      & 0.00                    & 0.00             & 0.00                     & 0.32              & 0.15                                              & 0.00                                        & 0.00                                              & 0.10                                        & 0.00                     & 0.00              & 0.18                                               & 0.26                                         & 0.13                 & \textbf{0.67}           & 0.42                  & 0.47             & 0.11                  & 0.23            \\
\begin{tabular}[c]{@{}l@{}}generalized \\ contraposition\end{tabular} & 0.00                    & 0.00             & 0.00                     & 0.16              & 0.15                                              & 0.00                                        & 0.33                                              & 0.21                                        & 0.00                     & 0.00              & 0.20                                               & 0.04                                         & 0.18                 & 0.27           & 0.43                  & 0.32             & 0.16                  & 0.13            \\
\begin{tabular}[c]{@{}l@{}}generalized \\ dilemma\end{tabular}        & 0.00                    & 0.00             & 0.00                     & 0.37              & \textbf{0.49}                                              & 0.00                                        & 0.15                                              & \textbf{0.56}                                        & 0.00                     & 0.00              & 0.34                                               & 0.18                                         & 0.26                 & 0.66  & 0.33                  & 0.53             & 0.20                  & 0.29            \\
\begin{tabular}[c]{@{}l@{}}generalized \\ modus ponens\end{tabular}   & 0.00                    & 0.00             & 0.00                     & \textbf{0.48}              & 0.28                                              & 0.05                                        & 0.32                                              & \textbf{0.56}                               & 0.00                     & 0.00              & \textbf{0.62}                                      & \textbf{0.61}                                & 0.18                 & 0.62           & 0.59                  & \textbf{0.74}             & 0.25                  & \textbf{0.38}            \\
\begin{tabular}[c]{@{}l@{}}generalized \\ modus tollens\end{tabular}  & 0.00                    & 0.00             & 0.00                     & 0.44              & 0.48                                              & 0.01                                        & 0.32                                              & 0.26                                        & 0.00                     & 0.00              & 0.39                                               & 0.17                                         & 0.11                 & 0.62           & \textbf{0.74}         & 0.73    & 0.25                  & 0.28            \\
\begin{tabular}[c]{@{}l@{}}hypothetical\\ syllogism 1\end{tabular}    & 0.00                    & 0.00             & 0.00                     & 0.40              & 0.45                                              & 0.00                                        & 0.31                                              & 0.39                                        & 0.00                     & 0.00              & 0.55                                               & 0.21                                         & 0.31                 & 0.54           & 0.68                  & 0.51             & \textbf{0.29}                  & 0.26            \\
\begin{tabular}[c]{@{}l@{}}hypothetical\\ syllogism 3\end{tabular}    & 0.00                    & 0.00             & 0.00                     & 0.29              & 0.13                                              & 0.00                                        & \textbf{0.41}                                              & 0.33                                        & 0.00                     & 0.00              & 0.17                                               & 0.27                                         & \textbf{0.39}                 & 0.65           & 0.70                  & 0.57             & 0.23                  & 0.26            \\ \bottomrule
\end{tabular}%
}
\caption{Reasoning accuracy results from Task 2 baseline models on the set of biologically factual argumentative texts (without synthetic data and without distractors).}
\label{tab:task2_ra_byscheme}
\end{table}

\begin{table}[]
\centering
\resizebox{0.5\columnwidth}{!}{%
\begin{tabular}{@{}lllllllll@{}}
\toprule
\multirow{2}{*}{Scheme}                                              & \multicolumn{2}{l}{Gemma-7b}     & \multicolumn{2}{l}{Gemma-7b-it}  & \multicolumn{2}{l}{\begin{tabular}[c]{@{}l@{}}Mistral-7B\\ Instruct-v0.2\end{tabular}} & \multicolumn{2}{l}{\begin{tabular}[c]{@{}l@{}}Mixtral-8x7B\\ Instruct-v0.1\end{tabular}} \\ \cmidrule(l){2-9} 
                                                                     & r               & p-val          & r               & p-val          & r                                          & p-val                                     & r                                           & p-val                                      \\ \midrule
\multicolumn{9}{l}{ZS}                                                                                                                                                                                                                                                                                                  \\ \midrule
\begin{tabular}[c]{@{}l@{}}disjunctive\\ syllogism\end{tabular}      & \textbf{-0.943} & \textbf{0.005} & 0.029           & 0.957          & -0.086                                     & 0.872                                     & 0.486                                       & 0.329                                      \\
\begin{tabular}[c]{@{}l@{}}generalized\\ contraposition\end{tabular} & \textbf{-0.943} & \textbf{0.005} & \textbf{-1.000} & \textbf{0.000} & \textbf{-0.829}                            & \textbf{0.042}                            & 0.429                                       & 0.397                                      \\
\begin{tabular}[c]{@{}l@{}}generalized\\ dilemma\end{tabular}        & \textbf{-1.000} & \textbf{0.000} & -0.086          & 0.872          & \textbf{1.000}                             & \textbf{0.000}                            & 0.314                                       & 0.544                                      \\
\begin{tabular}[c]{@{}l@{}}generalized\\ modus ponens\end{tabular}   & \textbf{-0.943} & \textbf{0.005} & \textbf{-1.000} & \textbf{0.000} & 0.543                                      & 0.266                                     & 0.771                                       & 0.072                                      \\
\begin{tabular}[c]{@{}l@{}}generalized\\ modus tollens\end{tabular}  & \textbf{-0.943} & \textbf{0.005} & \textbf{-0.829} & \textbf{0.042} & \textbf{0.829}                             & \textbf{0.042}                            & 0.771                                       & 0.072                                      \\
\begin{tabular}[c]{@{}l@{}}hypothetical\\ syllogism 1\end{tabular}   & \textbf{-1.000} & \textbf{0.000} & \textbf{-1.000} & \textbf{0.000} & 0.029                                      & 0.957                                     & -0.486                                      & 0.329                                      \\
\begin{tabular}[c]{@{}l@{}}hypothetical\\ syllogism 3\end{tabular}   & \textbf{-0.886} & \textbf{0.019} & \textbf{-0.943} & \textbf{0.005} & \textbf{-0.943}                            & \textbf{0.005}                            & 0.029                                       & 0.957                                      \\ \midrule
\multicolumn{9}{l}{FS}                                                                                                                                                                                                                                                                                                        \\ \midrule
\begin{tabular}[c]{@{}l@{}}disjunctive\\ syllogism\end{tabular}      & \textbf{-1.000}           & \textbf{0.000}          & -0.543 & 0.266 & \textbf{0.829}                             & \textbf{0.042}                            & -0.143                                      & 0.787                                      \\
\begin{tabular}[c]{@{}l@{}}generalized\\ contraposition\end{tabular} & -0.086  & 0.872 & \textbf{-0.943} & \textbf{0.005} & \textbf{-0.943}                            & \textbf{0.005}                            & \textbf{0.829}                                      & \textbf{0.042}                                      \\
\begin{tabular}[c]{@{}l@{}}generalized\\ dilemma\end{tabular}        & -0.657          & 0.156          & \textbf{-0.943}           & \textbf{0.005}          & \textbf{-0.829}                                     & \textbf{0.042}                                     & \textbf{0.943}                                      & \textbf{0.005}                                      \\
\begin{tabular}[c]{@{}l@{}}generalized\\ modus ponens\end{tabular}   & 0.371          & 0.468          & \textbf{-0.886} & \textbf{0.019} & 0.600                             & 0.208                            & \textbf{0.829}                              & \textbf{0.042}                             \\
\begin{tabular}[c]{@{}l@{}}generalized\\ modus tollens\end{tabular}  & 0.486           & 0.329          & \textbf{-1.000} & \textbf{0.000} & 0.371                                      & 0.468                                     & 0.257                                       & 0.623                                      \\
\begin{tabular}[c]{@{}l@{}}hypothetical\\ syllogism 1\end{tabular}   & -0.086           & 0.872          & \textbf{-0.943} & \textbf{0.005} & 0.314                                      & 0.544                                     & \textbf{0.829}                              & \textbf{0.042}                             \\
\begin{tabular}[c]{@{}l@{}}hypothetical\\ syllogism 3\end{tabular}   & \textbf{-1.000}          & \textbf{0.000}          & \textbf{-1.000} & \textbf{0.000} & \textbf{-0.943}                                     & \textbf{0.005}                                     & -0.486                             & 0.329                             \\ \bottomrule
\end{tabular}%
}
\caption{Values of the Spearman’s Ranked Correlation Coefficient (r) for Accuracy by Distractors and Syllogistic Scheme for evaluated Models in Task 2: Spearman's correlation coefficients (r) and p-values for the reasoning accuracy metric are shown across various levels of distractors and syllogistic schemes for each model. Negative r values reflect a decrease in accuracy with increasing distractor complexity. The highest correlation values for each scheme are highlighted in bold, indicating the models most affected by distractors.}
\label{tab:correlation_by_distractors_task2}
\end{table}

\end{document}